\documentclass[10pt,twocolumn,letterpaper]{article}
\usepackage{3dv}
\usepackage{times}
\usepackage{epsfig}
\usepackage{graphicx}
\usepackage{amsmath}
\usepackage{amssymb}

\usepackage{bm}

\usepackage{algpseudocode}
\usepackage{algorithm}

\usepackage{amsthm}
\usepackage{amsbsy}
\usepackage{booktabs}
\usepackage{enumitem}
\usepackage{mathtools}
\usepackage{subfigure}
\usepackage{multirow}
\usepackage{caption}
\usepackage{overpic}

\usepackage{xcolor}
\usepackage{balance}
\usepackage{tabularx}
\usepackage{makecell}

\usepackage[numbers,sort,compress]{natbib}

\usepackage[title]{appendix}

\usepackage{enumitem}
\setitemize{noitemsep,topsep=0pt,parsep=0pt,partopsep=0pt}

\DeclarePairedDelimiter{\norm}{\big\lVert}{\big\rVert}
\DeclarePairedDelimiter{\bignorm}{\Big\lVert}{\Big\rVert}
\DeclareMathOperator*{\argmin}{\arg\!\min}

\newlength{\oldparindent}
\newcommand{\modelname}{SkiRT\xspace}

\renewcommand{\paragraph}[1]{\vspace{5pt}{\noindent\textbf{#1}}}

\newcommand*{\affaddr}[1]{#1} 
\newcommand*{\affmark}[1][*]{\textsuperscript{#1}}
\newcommand*{\email}[1]{\small{\texttt{#1}}}

\newcommand{\Ti}{\hat{T}_i}

\usepackage[pagebackref=true,breaklinks=true,colorlinks,bookmarks=false]{hyperref}

\threedvfinalcopy % *** Uncomment this line for the final submission

\def\threedvPaperID{90} % *** Enter the 3DV Paper ID here

\begin{document}

%%%%%%%%% TITLE
\title{Neural Point-based Shape Modeling of Humans in Challenging Clothing}

\author{
Qianli Ma\affmark[1,2]\quad
Jinlong Yang\affmark[2] \quad
Michael J. Black\affmark[2] \quad
Siyu Tang\affmark[1]
\\
\affaddr{\affmark[1]ETH Z\"urich} \quad 
\affaddr{\affmark[2]Max Planck Institute for Intelligent Systems, T\"ubingen, Germany}\\
\email{\{qianli.ma, siyu.tang\}@inf.ethz.ch}, \quad \email{\{qma,jyang,black\}@tuebingen.mpg.de} 
}

\twocolumn[{%
\renewcommand\twocolumn[1][]{#1}%
\maketitle
\begin{center}
    \newcommand{\teaserwidth}{\textwidth}
    \vspace{-0.2in}
    \centerline{
        \includegraphics[width=\textwidth]{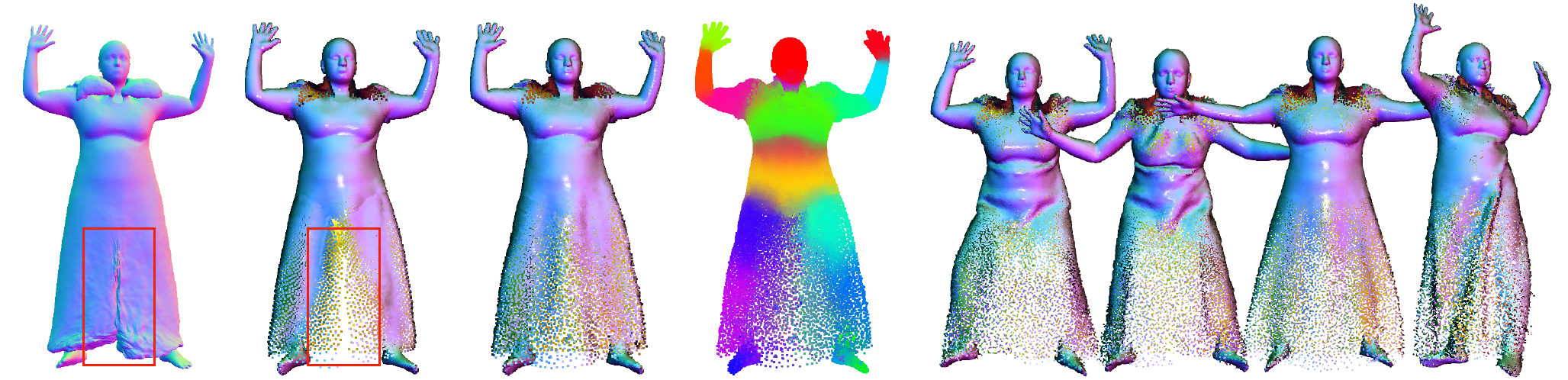}

    \put(-498,-10){\small \textbf{(a)} SCANimate~\cite{SCANimate:CVPR:2021} \quad\enspace \textbf{(b)} POP~\cite{POP:ICCV:2021} \qquad\quad \textbf{(c)} Ours \quad \textbf{(d)} Predicted LBS Weights \qquad\qquad\qquad \textbf{(e)} Ours, animated}
    }
    \captionof{figure}{\small\textbf{Learning clothed avatars with \modelname (Ours)}.
    When applied to challenging clothing types such as skirts and dresses, 
    existing methods for learning clothed human shape often \textbf{(a)} erroneously generate pant-like structures \cite{SCANimate:CVPR:2021}, or \textbf{(b)} suffer from inhomogeneous point density and split-like artifacts in the predicted shape \cite{POP:ICCV:2021}.
    Our point-based approach \textbf{(c)} addresses these issues by predicting the LBS weights relating the clothed surface to the body \textbf{(d)} and using a novel coarse shape representation in the posed space.
    \modelname achieves state-of-the-art modeling accuracy for pose-dependent shapes of humans in diverse types of clothing.
    Results from the point-based methods here are rendered using a plain point renderer to highlight the difference between each method's ``raw'' outputs.}
    \label{fig:teaser}
\end{center}%
}]

\maketitle
\thispagestyle{empty}

%%%%%%%%% ABSTRACT
\begin{abstract}
Parametric 3D body models like SMPL only represent minimally-clothed people and are hard to extend to clothing because they have a fixed mesh topology and resolution.
To address these limitations, recent work uses implicit surfaces or point clouds to model clothed bodies.
While not limited by topology, such methods still struggle to model clothing that deviates significantly from the body, such as skirts and dresses.
This is because they rely on the body to canonicalize the clothed surface by reposing it to a reference shape.
Unfortunately, this process is poorly defined when clothing is far from the body.
Additionally, they use linear blend skinning to pose the body and the skinning weights are tied to the underlying body parts.
In contrast, we model the clothing deformation in a local coordinate space without canonicalization.
We also relax the skinning weights to let multiple body parts influence the surface.
Specifically, we extend point-based methods with a coarse stage, that replaces canonicalization with a learned pose-independent ``coarse shape'' that can capture the rough surface geometry of clothing like skirts. 
We then refine this using a network that infers the linear blend skinning weights and pose dependent displacements from the coarse representation. 
The approach works well for garments that both conform to, and deviate from, the body.
We demonstrate the usefulness of our approach by learning person-specific avatars from examples and then show how they can be animated in new poses and motions. 
We also show that the method can learn directly from raw scans with missing data, greatly simplifying the process of creating realistic avatars. 
Code is available for research purposes at {\small\url{https://qianlim.github.io/SkiRT}}.
\end{abstract}
\section{Introduction}
A wide range of applications are driving recent interest in creating animatable 3D human avatars from data.
Given a set of 3D scans or meshes of a person performing various poses, the goal is to create a model of the clothed body that can be controlled by the body pose, with the clothing deforming naturally as the body moves. 
Such technology will greatly simplify the creation of avatars for immersive experiences in teleconferences, gaming, and virtual try-on, to name a few.
Traditional solutions to this task typically require an artist-designed clothing template, a step that registers the template to data, and a process to associate the registered data to a controllable bone structure. 
These all involve expert knowledge and manual effort. 

While registering a mesh template to a clothed body scan is challenging and  error-prone, there are tractable solutions to estimate the minimally-clothed body pose and shape that lies under clothing~\cite{Zhang_2017_CVPR, AGORA:CVPR:21,bhatnagar2020loopreg,bhatnagar2020ipnet,PTF:CVPR:2021}, i.e., to fit a parametric human body model \cite{loper2015smpl, pavlakos2019expressive, xu2020ghum, joo2018total} to clothed scans.
This gives rise to a promising alternative to the complex traditional pipeline.
Recent methods build shape models of \textit{clothed} humans using clothed scans and the corresponding minimally-clothed bodies, thereby side-stepping the need for per-garment mesh templates \cite{patel20tailornet, bertiche2020cloth3d,santesteban2019}.

%%%% 
To go beyond  minimally-clothed body models such as SMPL \cite{loper2015smpl} to capture the complex shape of clothed bodies, recent attempts either displace the vertices of the body template mesh \cite{CAPE:CVPR:20, burov2021dsfn,tiwari20sizer,Neophytou2014layered} or learn a neural implicit field (without relying on templates) that captures the clothing shape conditioned on the body pose parameters \cite{SCANimate:CVPR:2021, chen2021snarf, tiwari21neuralgif,MetaAvatar:arXiv:21}.
While these methods show promising results on T-shirts, pants, or even closed jackets, they exhibit limitations for clothing types that are topologically different from the body, like dresses and skirts. 
To date, these remain a major challenge for both tempate-based and template-free models of clothed humans.

%%%% 
One of the key problems of these existing formulations is that they require explicit \textit{canonicalization}.
Some work assumes the canonicalized data is given~\cite{CAPE:CVPR:20, palafox2021npm}, whereas the recent trend is to learn a backward transformation field to ``unpose'' the data to a consistent canonical pose~\cite{MetaAvatar:arXiv:21, SCANimate:CVPR:2021, chen2021snarf}. 
On a high-level, canonicalization is a process that transforms all points on the posed, clothed body surface to a pre-defined pose, using the bone transformations derived from the underlying body.
Here, a hidden assumption is that the unposing of clothing can be driven simply by the unposing of the minimally-clothed body. 
While this is true for many clothing types that lie close to the body (such as T-shirts and pants), the assumption often fails for garments that differ greatly from the body shape and topology, such as skirts and dresses. The canonicalization step for a skirt is ill-defined, and canonicalizing such surfaces with existing body-based approaches often leads to artifacts, as we demonstrate in Sec.~\ref{sec:analyze}; see Fig.~\ref{fig:scanimate_failure}.

%%%%
In contrast to canonicalization methods, shape deformation can be modeled in local, posed, coordinates; e.g.,
recent point-based clothed human models \cite{SCALE:CVPR:2021, POP:ICCV:2021} directly model shape in the \textit{posed} space. 
For a query point on the body surface, a clothing displacement vector is predicted in a pre-defined local coordinate system anchored to that point.
These methods represent the pose-dependent shapes of skirts by simple offsets from the unclothed body model.
However, as the definition of the local coordinate frame relies on the linear blend skinning (LBS) weights of SMPL (Sec.~\ref{sec:analyze}), these methods are tied to body part transformations, resulting in clear surface discontinuities on the skirts and dresses; see Fig.~\ref{fig:teaser}.

Here, we follow the point-based approach but further equip it with a few key innovations that combine the traditional pipeline with implicit surface-based body modeling techniques. 
Specifically, we extend the point-based method, POP~\cite{POP:ICCV:2021}, with a learned, pose-\textit{in}dependent ``coarse shape'' that captures the rough surface geometry of clothing. 
The coarse shape serves a similar role as the clothing template in the traditional pipeline. 
Instead of using a manually crafted mesh, however, our point-based coarse geometry is learned from data, is not tied to specific mesh topology, and is hence flexible in representing various clothing types.
We refine this coarse shape using a neural network that predicts pose-dependent displacements  from the coarse representation, in their respective local coordinates.
Differing from prior work~\cite{POP:ICCV:2021}, the transformations from the local to the world coordinates are not taken from the underlying body model, but \textit{predicted} by a neural network. 
Additionally, we introduce an adaptive regularization technique to train the pipeline, which encourages a more uniform spatial distribution of the generated point clouds.
As we show in the experiments, these innovations lead to improved modeling accuracy and, more importantly, are essential to solving the typical surface discontinuity problems found in previous work.

These key components enable our new method, \modelname (\textit{Skinned Refined Template-free}), which creates pose-dependent shape models of clothed humans from 3D scans.
We evaluate \modelname on diverse clothing types, with a special focus on skirts and dresses of different lengths, tightness and styles; of course, \modelname also models simpler clothing types.
\modelname tackles the long-standing difficulty of modeling challenging clothing types such as skirts and dresses, 
outperforming state-of-the-art methods from both the point-based and implicit surface-based families on the pose-dependent clothing modeling task.
Furthermore, with an extensive ablation study, we carefully analyze the key components in our pipeline, providing insights for future point-based models of clothed human body geometry.

\section{Related Work}

In this section we review previous work that captures and models 3D clothing with data-driven learning approaches.

\paragraph{Template-based clothed human modeling.}
Existing graphics software and, in particular, 3D clothing simulation, uses 3D meshes to represent human bodies and clothing.
Physics-based cloth simulation has been widely used to train data-driven clothing models~\cite{de2010stable, xu2014sensitivity, bertiche2020cloth3d, santesteban2019, vidaurre2020fully, patel20tailornet, jiang2020bcnet,santesteban2022snug}, and 
mesh-based models of minimally-clothed humans are common
 \cite{anguelov2005scape, loper2015smpl, pavlakos2019expressive, xu2020ghum}.
 Mesh-based body models have been a foundation for capturing and representing clothing deviation from the the body
\cite{guan2012drape, vlasic2008articulated, Neophytou2014layered, yang2018analyzing, pons2017clothcap, CAPE:CVPR:20, lahner2018deepwrinkles, zhu2020deep}. 
While deep learning methods can be used to model clothing deviation on meshes \cite{ranjan2018generatingcoma, gong2019spiralnet++, verma2018feastnet}, these methods remain limited when it comes to skirts and dresses.
Unlike many pants and tops, skirts and dresses differ greatly from the body surface, making it infeasible to deform a common body template mesh to fit the data.
To capture skirts and dresses, some methods exploit customized template meshes~\cite{guan2012drape, pons2017clothcap, vlasic2008articulated, habermann2020deepcap, zhu2020deep, zhu2022registering}. 
With such an approach, one must manually design templates for a wide variety of clothing styles, making this approach impractical.
In contrast, our method is ``template-free'' in that it 
replaces the per-garment mesh templates with data-driven coarse shapes that serve a similar role. This obviates the expensive manual labor and can, in principle, model arbitrary skirt styles.

\paragraph{Template-free clothed human modeling.} To deal with the problems of template-based approaches, recent work goes beyond meshes and exploits new representations that do not require a per-garment template. 
A popular and promising approach uses deep implicit shapes \cite{mescheder2019occupancy, park2019deepsdf, chen2019imnet, chibane2020ndf} to capture~\cite{bhatnagar2020ipnet, saito2019pifu,saito2020pifuhd,xiu2022icon,xu2021hnerf,dong2022pina,ARAH:ECCV:2022} and model clothed humans ~\cite{SCANimate:CVPR:2021, chen2021snarf,xu2022gdna, MetaAvatar:arXiv:21, chibane2020implicit, tiwari21neuralgif, palafox2021npm,palafox2021spams}. 
Although such representations adapt to varied topology, these methods \cite{SCANimate:CVPR:2021, chen2021snarf, MetaAvatar:arXiv:21, palafox2021npm} do not perform well when modeling skirts, especially wide and long skirts. 
The problem originates from the need to canonicalize the training data as detailed in Sec.~\ref{sec:analyze}. 
In contrast, point clouds provide an alternative representation.
Ma et al.~\cite{SCALE:CVPR:2021} successfully model clothing or clothed humans with point clouds, and  
further demonstrate in \cite{POP:ICCV:2021} that their approach can be extended to train a single model of various clothing styles. 
While promising, dense point clouds can have trouble with skirts, producing a discontinuous representation between the legs. The reason is that these methods are still tied to the topology of the unclothed body, as we analyze in Sec.~\ref{sec:analyze}.
Our work uses a ``coarse stage" with a local coordinate space to avoid the issues in previous work. 
It also uses a network to predict more reasonable skinning weights for the skirt points, minimizing the discontinuities seen with previous work.
In a similar spirit,  concurrent work~\cite{lin2022fite} also enhances the point-based clothing representation with a coarse template; but~\cite{lin2022fite} uses diffusion to acquire clothing skinning weights, whereas our method learns the skinning weights from data. 
\section{Analyzing the Challenges of Skirts}\label{sec:analyze}
Given a clothed scan with known body pose, many methods transform the clothed scan to a common canonical pose. 
This helps factor out the articulation-related deformations in the clothed body shape, reducing data variation and facilitating model learning. 
A premise for canonicalization to work is that the clothing, when dressed on the body, behaves similarly to the unclothed body under different poses. 
Many tops and pants mostly satisfy this assumption, as evidenced by the success of recent canonicalization-based models on these garment types \cite{pons2017clothcap,CAPE:CVPR:20,SCANimate:CVPR:2021,chen2021snarf,MetaAvatar:arXiv:21}.
However, the surface deformation of skirts and dresses does not directly follow the body articulation.
Consequently, extra care needs to be taken to canonicalize and animate such clothing \cite{pons2017clothcap}, e.g., by introducing extra/virtual bones~\cite{liu2019neuroskinning,pan2022predicting}.

\begin{figure}[tb]
    \centering
    \includegraphics[width=0.9\linewidth]{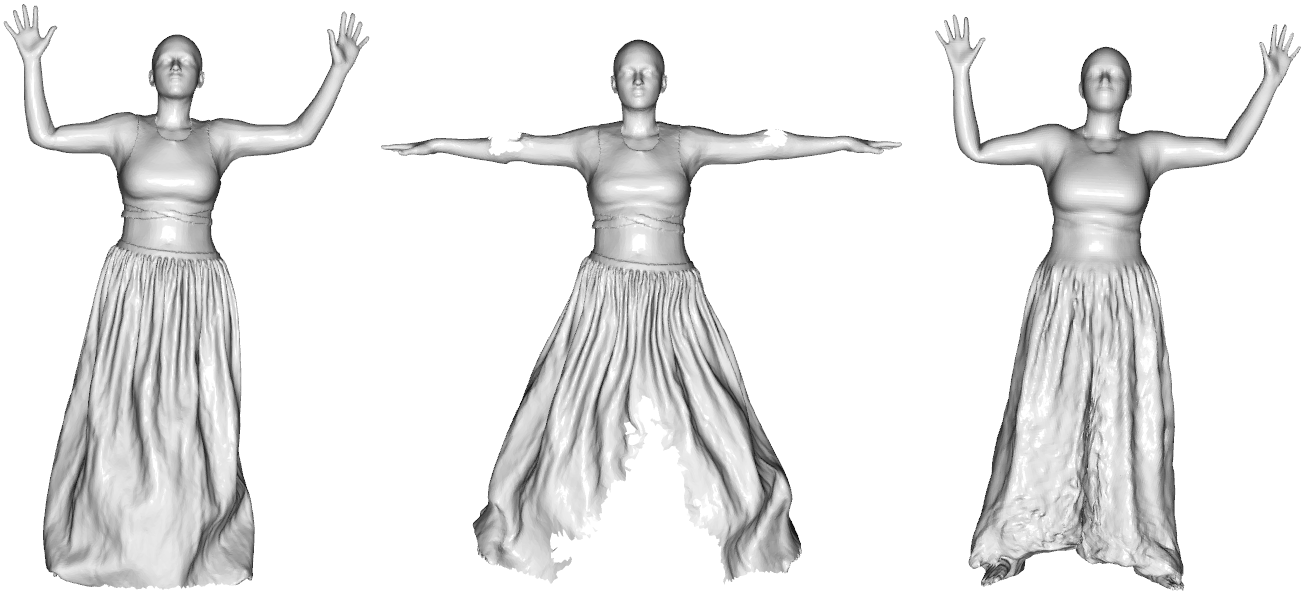}
    \put(-224,5){(a)}
    \put(-153,5){(b)}
    \put(-73,5){(c)}
    \caption{\small Typical failure case due to canonicalization of skirts using the learned inverse-LBS methods. Result produced by \cite{SCANimate:CVPR:2021} on a training example. Given a raw scan (a), canonicalization often tears or creases the skirt (b), leading to areas in canonical space without good supervision. A neural implicit function trained with such canonicalized shapes yields visible artifacts (c).}
    \label{fig:scanimate_failure}
\end{figure}

Figure~\ref{fig:scanimate_failure} shows a typical failure mode of a state-of-the-art canonicalization method, SCANimate~\cite{SCANimate:CVPR:2021}, on a dress.
Here, the clothing-body association is predicted by a neural network, which is learned from data. Theoretically, the network should be able to learn an optimal canonical skirt shape such that it matches the data when transformed back to the posed space.
However, in practice, it is difficult to avoid over-stretching and splitting the raw scans.
When the neural network is trained to fill the gap without supervision, reposing the canonical representation produces creases.
As a result, even on training data (as in Fig.~\ref{fig:scanimate_failure}), the learned reconstructed skirt shape typically has a clear crease, with the resulting garment looking more like wide pants.

\begin{figure}[tb]
    \centering
    \includegraphics[width=0.85\linewidth]{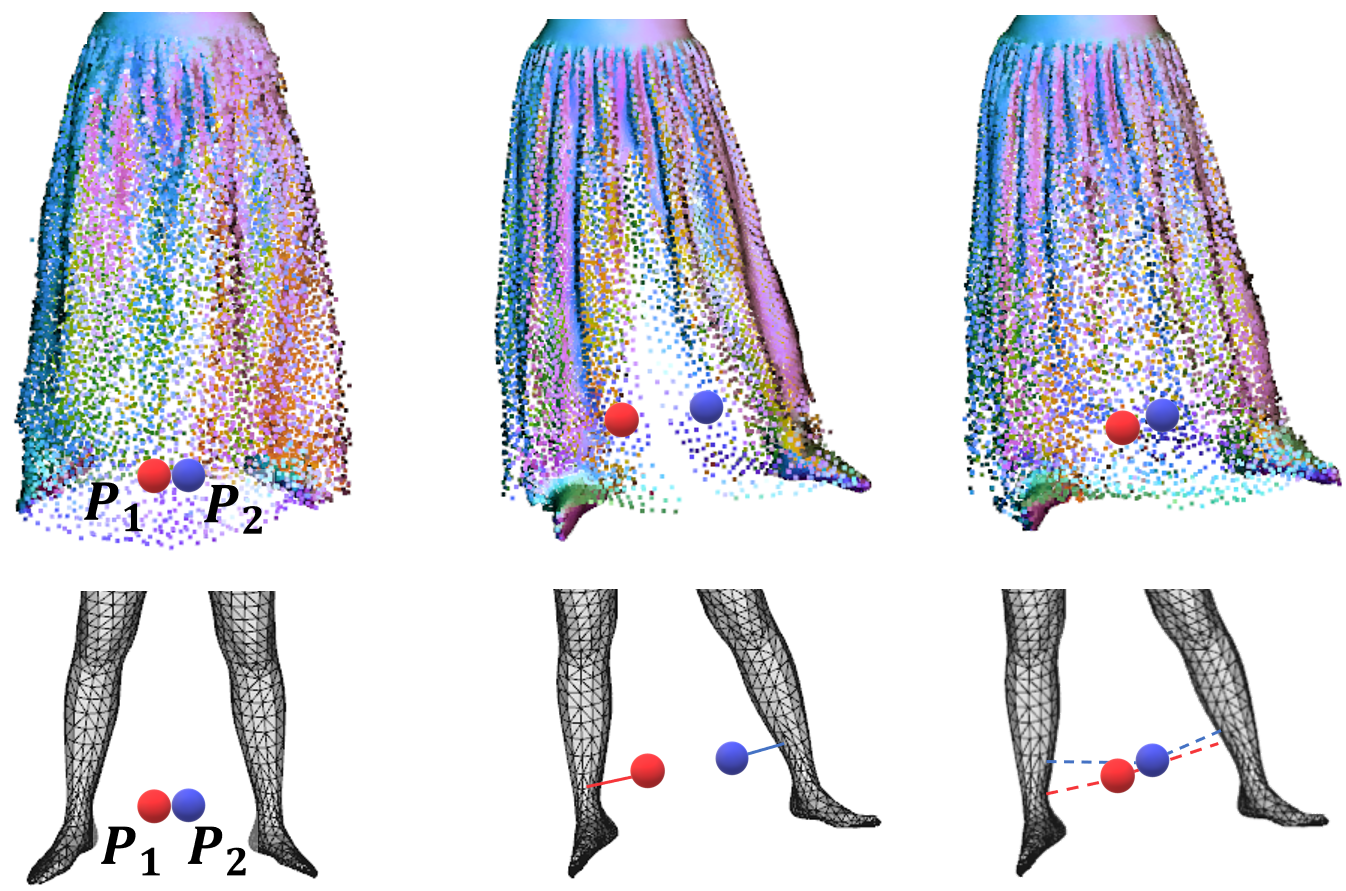}
    \put(-210,120){(a)}
    \put(-142,120){(b)}
    \put(-70,120){(c)}
    \vspace{-3pt}
    \caption{\small\textbf{Failure mode} of the SoTA point-based method on a skirt. (a) Points $P_1, P_2$ denote two neighboring points on the skirt surface, displaced from the left and right leg respectively. When the pose changes, (b), previous work \cite{POP:ICCV:2021} skins these according to the body point skinning weights, resulting in a ``hard'' association to only the respective body part (the solid colored lines). This leads to a discontinuous skirt surface in the new pose. (c) Our approach predicts blended skinning weights for each point (the dashed colored lines) leading to a more even distribution of points.}
    \label{fig:2d_illustr}
\end{figure}

The recent point-based clothed human model, POP \cite{POP:ICCV:2021}, side-steps explicit canonicalization and models the clothed body shapes in the posed space using local coordinates.
While showing the potential of capturing clothing that differs from the body topology, POP typically exhibits a failure mode 
when representing skirts: the skirt surface is often torn apart in the region between the legs, as shown in Figs.~\ref{fig:teaser} and \ref{fig:2d_illustr}. 
One of the key reasons is that the local coordinates are defined based on the unclothed body shape and topology, as we analyze below.

In a nutshell, the POP model densely queries locations on the body surface and generates a clothing displacement vector at each query. 
For a query point $b_i\in \mathbb{R}^3$ on the \textit{posed} body, the clothing displacement $\hat{r}_i \in \mathbb{R}^3$ is predicted in its associated local coordinate system. The local coordinate frame is a Cartesian coordinate system with $b_i$ being the origin. 
The clothing point's location in world coordinates is then given by:
\begin{equation}\label{eq:pop}
    x_i = \Ti \cdot \hat{r}_i + b_i.
\end{equation}
In POP, the transformations $\Ti$ between the local and world coordinate systems are \textit{pre-computed} using the joint transformations of the SMPL model:
\begin{equation}\label{eq:lbsw}
    \Ti = \sum_{j=1}^{J} w_{j,\textrm{SMPL}}^{(i)} T_j,
\end{equation}
where each $T_j$ is the $4\times4$ rotation-translation transformation matrix of the body's $j$-th joint from the canonical-pose space to the posed space, $J$ is the number of joints, and $w_{j,\textrm{SMPL}}^{(i)}$ is SMPL's linear blend skinning (LBS) weight that associates $b_i$ to the $j$-th joint, s.t. $\sum_{j=1}^J w_{j,\textrm{SMPL}}^{(i)}= 1$.

Notice that $\Ti$ in fact also yields $b_i$ (on the posed body) from its corresponding location $\hat{b}_i$ in the canonical pose\footnote{With a slight abuse of notation, $b_i$ can represent the homogeneous coordinates when necessary.}: $b_i = \Ti \cdot \hat{b}_i$.
It then becomes clear that the POP pipeline is equivalent to building the model in the canonical space:
\begin{equation}\label{eq:pop_cano}
    x_i = \Ti \cdot \hat{x}_i, 
\end{equation}
with $\hat{x}_i = \hat{r}_i + \hat{b}_i$ being the clothed body in the canonical pose. 
While the predictions are made in the canonical space, POP does not explicitly canonicalize the scan data, and uses the original posed data as supervision during training. This bypasses the drawbacks of the explicit canonicalization as earlier discussed. 

However, the pre-computed local coordinate transformations are problematic for skirts. 
The points on the skirt surface are displaced from \textit{either} the left or right leg. 
Consider two neighboring points, $P_1$ and $P_2$, from the between-leg region on the skirt surface in Fig.~\ref{fig:2d_illustr}. 
$P_1$ and $P_2$ are displaced from the right and left leg, respectively.
With POP, the LBS weights for $P_1, P_2$ are effectively drawn from one leg or the other; i.e.~
$w^{P_1}_\textit{right leg} \approx 1$ and $w^{P_2}_\textit{left leg} \approx 1$, while other LBS weights from other bones are approximately 0.
When the leg pose changes, $P_1$ and $P_2$ thus follow the rigid transformation of \textit{only} their respective leg.
When the legs move apart, so do the points on the skirt,
Fig.~\ref{fig:2d_illustr}(b).
Although POP's predicted non-rigid pose-dependent displacements can, in theory, compensate for the discrepancy of the rigid transforms, in our experiments, we find this insufficient.
Specifically, the proximity of points on the skirt varies with pose.
This results in a gap in the 3D representation that splits the skirt into two parts. 
In fact, the local transformations, as defined in Eq.~\eqref{eq:lbsw}, result in a discontinuity of the displacement field on the skirt surface. We show this with an experiment in Sec.~\ref{sec:exp_sota_comparison}.

%%%%%%%%%%%%%%%%%%%%%%%%%%%%%%%%%%%%%%%%%%%%%%%%%%%%%%%%
\section{Method}
\begin{figure}[tb]
    \centering
    \includegraphics[width=\linewidth]{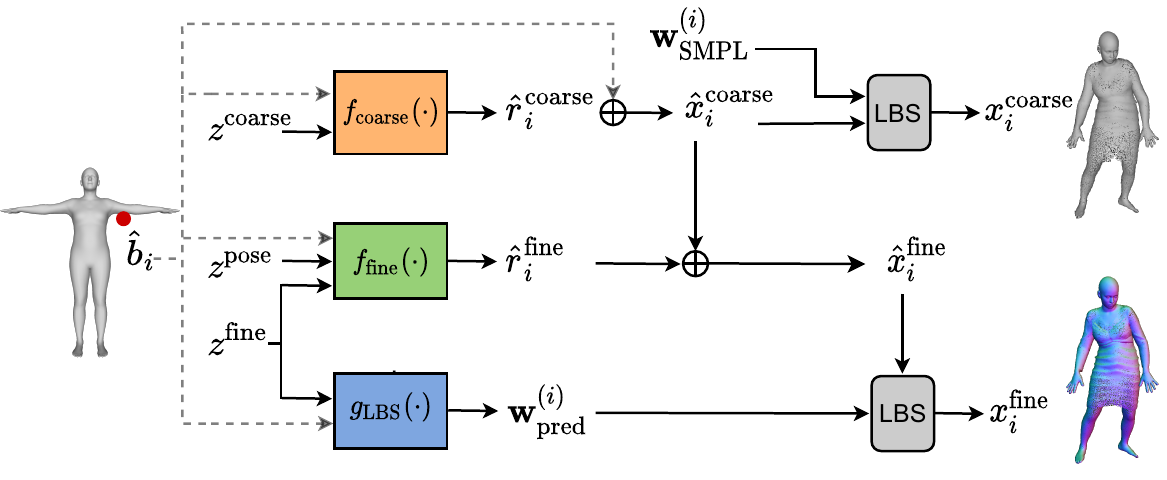}
    \caption{\small \textbf{Method Overview.}
    Given a point $\hat{b}_i$ on the body surface in canonical pose, we use it to query the three MLP networks that predict the pose-independent coarse clothing shape $\hat{r}^\textrm{coarse}$, the pose-dependent detailed clothing offsets $\hat{r}^\textrm{fine}$ (Sec.~\ref{sec:coarse2fine}), and the LBS weights on the clothing $\mathbf{w}^{(i)}_\textrm{pred}$ (Sec.~\ref{sec:pred_lbsw}), respectively. The learned LBS weights are used to transform the canonical-space prediction $\hat{x}_i^\textrm{fine}$ to the posed space $x_i^\textrm{fine}$.
    }
    \label{fig:method_overview}
\end{figure}

\paragraph{Overview.} To cope with the challenging clothing types such as skirts and dresses, we 
push the limit of the local coordinate-based point representation by introducing a series of new components.
The training of our method takes a set of scans (point clouds) that capture the shape of a person wearing the same outfit in varying poses. For each scan, we fit a SMPL body that captures the body shape and pose under clothing. We first train a coarse-stage neural network that takes in the SMPL bodies and predicts a pose-\textit{in}dependent shape of the clothed body (Sec.~\ref{sec:coarse2fine}). 
We then refine this coarse shape with another neural network that predicts a pose-dependent, detailed clothing displacement field in the local coordinates. The fine stage also predicts a field of transformations of these local coordinates in the form of LBS weights (Sec.~\ref{sec:pred_lbsw}). 
The displacements are then added to the coarse shape and posed using the predicted transformations, yielding the final clothed body prediction.
We additionally introduce an adaptive regularization algorithm in Sec.~\ref{sec:adaptive_loss} to facilitate training the pipeline.
An overview of our method is illustrated in Fig.~\ref{fig:method_overview}.

%%%%%%%%%
\subsection{Coarse-to-fine Prediction} \label{sec:coarse2fine}
Based on the analysis in Sec.~\ref{sec:analyze}, an immediate remedy is to associate the points on a skirt to multiple body parts, instead of a ``hard'' association to only one part. 
Taking Fig.~\ref{fig:2d_illustr} again as an intuitive example: if both points $P_1, P_2$ are associated with both legs, they will undergo similar rigid transformations as the legs move, hence maintaining their proximity, Fig.~\ref{fig:2d_illustr}(c). 
This amounts to ``spreading'' the LBS weights to all the leg joints instead of tightly associating them with one, such that the LBS weights on the skirt surface vary smoothly in space. 
The remaining question is, how to get such smoothly transitioning LBS weights on the clothed body surface? 
Our answer is to first learn a pose-independent coarse shape and then use it to query a pre-diffused smooth LBS weight field derived from SMPL. 

\paragraph{Pose-independent coarse shape.} We first introduce a neural network that predicts the coarse shape of clothing, represented as residual vectors $ \hat{r}_i^\textrm{coarse}$ from the body:
\begin{equation} \label{eq:coarse_mlp}
    \hat{r}_i^\textrm{coarse} = f_\textrm{coarse}(\hat{b}_i, z^\textrm{coarse}):\mathbb{R}^3 \times \mathbb{R}^{\mathbb{Z}^\textrm{coarse}} \rightarrow \mathbb{R}^{3},
\end{equation}
where $f_\textrm{coarse}(\cdot)$ is a multi-layer-preceptron (MLP), $\hat{b}_i$ is a query point on the neutrally-posed body, and $z^\textrm{coarse}$ is a \textit{global} geometry code. 
For the coarse stage, we use the LBS from the SMPL model to bring the predictions back to the posed space, Eq.~\eqref{eq:pop}.
By querying over the entire body surface, a coarse shape of the clothed body can be constructed.

Since the coarse MLP does not take in any body pose information, the output is pose-independent. Training it on all data with varied poses results in a coarse clothed body shape that minimizes the discrepancy to all examples. 
Intuitively this facilitates the subsequent fine stage, where the network then only needs to predict a small pose-dependent residual on the coarse base shape.
The learned coarse shape takes the place of an artist-designed garment template in a traditional graphics pipeline, but the flexible point-based representation enables our method to be applied to different clothing types without manually defining garment templates.

%%%%%%%%%
\paragraph{Pose-dependent fine shape.} For the fine stage, we follow the design of POP~\cite{POP:ICCV:2021} and train another MLP, $f_\textrm{fine}(\cdot)$, that takes in a local geometry descriptor $z_i^\textrm{fine}$, a local pose descriptor $z_i^\textrm{pose}$, and outputs a 
clothing residual in the local coordinates:
\begin{equation} \label{eq:fine_mlp}
    \hat{r}_i^\textrm{fine} = f_\textrm{fine}(\hat{b}_i, z_i^\textrm{fine}, z_i^\textrm{pose}): \mathbb{R}^3 \times \mathbb{R}^{\mathbb{Z}^\textrm{fine}} \times \mathbb{R}^{\mathbb{Z}^\textrm{pose}}\rightarrow \mathbb{R}^{3},
\end{equation}
where the subscript $i$ of the descriptors denotes that they are local to each query point. 
Following POP, the geometry features $z_i^\textrm{coarse}$ and $z_i^\textrm{fine}$ are learned in an auto-decoding fashion.
But, unlike POP, the predicted residuals are now added to the coarse clothed body shape, instead of the unclothed body model. 
The $f_\textrm{fine}(\cdot)$ MLP branches out another head at the final layer to predict the normal vector in the local coordindates. More details of the local descriptors are provided in Appendix~\ref{sec:appendix:skirt_archi}.

%%%%%%%%%
\paragraph{Pre-diffused LBS weight field.}
With the learned clothed coarse body shape, we are now able to define smoothly-varying LBS weights for the coarse shape. 
To do so, we first diffuse SMPL's LBS weights to $\mathbb{R}^3$ using nearest neighbor assignment as in LoopReg~\cite{bhatnagar2020loopreg}. 
Such a pre-diffused LBS weight field has many undesirable discontinuities in space (see Appendix~\ref{sec:appendix:lbs_field}.).
We then run an optimization to smooth the field. To do so, we sample a regular grid in the diffused space, and minimize the discrepancy between each grid point's LBS weight and the average weight of its 1-ring neighbors; this is analogous to applying Laplacian smoothing to the LBS weights. 
The optimization results in a LBS weight field that has a smoother spatial variation. 
Finally, we use the points from the coarse shape to query this field, obtaining smoothly-varing LBS weights on the clothed body ``template".
More details about our LBS smoothing algorithm can be found in the Appendix~\ref{sec:appendix:lbs_field}.

%%%%%%%%%
\subsection{Predicting Local Transformations} \label{sec:pred_lbsw}
Although the optimized LBS weight field in theory addresses the single association problem, the optimization is based on heuristics and it is unclear if these pre-processed LBS weights are the optimal choice for every clothing type. 
Therefore, we train another neural network, $g_\textrm{LBS}(\cdot)$, to predict the LBS weights, and use the pre-diffused field for regularization:
\begin{equation} \label{eq:lbs_net}
  \mathbf{w}_\textrm{pred}^{(i)} = g_\textrm{LBS}(\hat{b}_i, z_i^\textrm{fine}): \mathbb{R}^3 \times \mathbb{R}^{\mathbb{Z}^\textrm{fine}}\rightarrow \mathbb{R}^{J},
\end{equation}
with $\mathbf{w}_\textrm{pred}^{(i)}=\{w_{j,\textrm{pred}}^{(i)}\}_{j=1}^J$ being the set of predicted LBS weights at the query location $\hat{b}_i$. Note that the LBS prediction network is conditioned on the clothing geometry features but not the pose features,
following the common design principle that the LBS weights are outfit-specific but pose-independent.

Using the predicted LBS weights, we modify Eq.~\eqref{eq:lbsw} as:
\begin{equation}\label{eq:pred_lbsw}
    \Ti' = \sum_{j=1}^{J} w_{j,\textrm{pred}}^{(i)} T_j.
\end{equation}
The clothing residuals and surface normals predicted from the fine stage, $\hat{r}_i^\textrm{fine}$, are now transformed with the predicted local transformations $\Ti'$.

%%%%%%%%%
\subsection{Point-adaptive Upsampling and Regularization}\label{sec:adaptive_loss}
In POP, when the skirt ``splits'', the points becomes exceptionally sparse on the region between the legs, Fig.~\ref{fig:2d_illustr}(b).
We propose an algorithm to increase the point density for such regions, aiming to eliminate such artifacts in a simple post-processing step.
To identify the points in the low-density region, we compute the average distance of each point to its k-nearest neighbors (kNN, we use k=5). 
This measures how isolated each point is from its neighbors, giving a measure of local point density. 
We then again query the underlying body surface using the kNN radius as  guidance, and assign a higher sampling probability to the regions with  low point density. 

As we show in the experiments, this simple post-process can make the point distribution more uniform and visually reduce the ``split'' artifact, but unfortunately does not eliminate it entirely. 
This indicates that the learned neural field for the skirt surface is still discontinuous.

To address this, we introduce the point-adaptive idea in training. For points that have a large kNN radius, we set a smaller weight $\lambda_i^\textrm{adapt}$ for the regularization on the norm of their displacements (see Eq.~\eqref{eq:rgl}). 
Intuitively, with a uniform sampling on the body surface, the displaced clothing points with a large kNN radius are more likely to be far from the body. Consequently, the clothing deformations there receive less influence from the body articulations, hence the non-rigid deformations (displacements) can have more freedom. 
As we show in the experiments, the point-adaptive regularization effectively improves the uniformity of the generated point clouds. 
Further details of the point-adaptive sampling are provided in the Appendix~\ref{sec:appendix:adaptive_sample}.

%%%%%%%%%
\subsection{Training and Inference}\label{sec:training}
Our pipeline is trained with a two-stage regime. The coarse shape network is trained first such that it provides a base shape for the fine stage training. It also serves as a medium to acquire the pre-diffused LBS weights for regularizing the LBS prediction. The pose-dependent fine shape network and the LBS weight prediction networks are trained together end-to-end. 

Since the coarse base shape does not vary with pose, when dealing with unseen poses at test-time, we only need to run the fine-stage inference given the query pose. 

\paragraph{Loss Functions.}
Both the coarse and fine stages are trained with a weighted sum of the Chamfer Distance $\mathcal{L}_\textrm{CD}$, a normal consistency loss $\mathcal{L}_\textrm{n}$, and a regularization term $\mathcal{L}_\textrm{rgl}$ on the norm of the predicted displacements. 
The standard Chamfer and normal losses follow the formulation in \cite{POP:ICCV:2021} and we defer their description in the Appendix~\ref{sec:appendix:skirt_archi}.

The predicted displacements $\hat{r}^\textrm{coarse}$ and $\hat{r}^\textrm{fine}$ are regularized by their L2-norm, weighted by the point-adaptive regularization strength $\lambda_i^\textrm{adapt}$ as discussed in Sec.~\ref{sec:adaptive_loss}:
\begin{equation}\label{eq:rgl}
\mathcal{L}_\textrm{rgl} = \frac{1}{|\mathbf{X}|}\sum_{i=1}^{|\mathbf{X}|}\lambda_i^\textrm{adapt}\norm{\hat{r}_i}_2^2.
\end{equation}

For the fine stage specifically, we use two regularization terms to facilitate the learning of the LBS weight prediction. The first one is a direct L1 regularization using the ground truth LBS weights from the underlying body:
\begin{equation}\label{eq:lbsw_l1}
\mathcal{L}_\textrm{LBS} = \frac{1}{|\mathbf{X}|}\sum_{i=1}^{|\mathbf{X}|}\norm{\mathbf{w}^{(i)}_\textrm{pred} - \mathbf{w}^{(i)\prime}_\textrm{SMPL}}_1,
\end{equation}
where $\mathbf{w}^{(i)\prime}_\textrm{SMPL}$ is the pre-diffused and smoothed SMPL LBS weights as discribed in Sec.~\ref{sec:coarse2fine}.

Additionally we introduce a reprojection loss as follows:
\begin{equation}
    \label{eq:reproj}
    \mathcal{L}_\textrm{reproj} = \frac{1}{|\mathbf{X}|}\sum_{i=1}^{|\mathbf{X}|}\norm{
    T_i'\hat{b}_i - b_i^{(gt)}}_2,
\end{equation}
where $\hat{b}_i$ is a sampled point on the body in canonical pose, $T_i'$ is the local transform derived from the predicted LBS weights (Eq.~\eqref{eq:pred_lbsw}), and $b_i^{(gt)}$ is the corresponding point on the ground truth body surface. That is, we use the learned transformations to reproject the canonical-posed body to the pose space, such that it matches the ground truth posed body.  
Similar to $\mathcal{L}_\textrm{LBS}$, this term also penalizes the learned LBS weights for deviating too far from those of the body.

Further details on the model architecture and training hyperparameters are provided in the Appendix~\ref{sec:appendix:skirt_archi}.
\section{Experimental Evaluation}

\begin{table*}[tb]
\centering
\caption{\label{table:quant} \small Quantitative comparison with baselines and the ablated versions of our method on the ReSynth~\cite{POP:ICCV:2021} dataset. CD: Chamfer Distance ($\times 10^{-4} m^2$); NML: L1 discrepancy between the predicted and ground truth unit normals ($\times 10^{-1}$). Both the lower the better. The best results are shown in bold.
The upper section contains subjects wearing skirt/dress outfits; the lower section contains subjects wearing non-skirt loose clothing such as jackets (carla-004, eric-035) and wide T-shirts with pants (alexandra-006).
}
\begin{tabular}{@{}rrc|cc|cc|cc|cc|cc@{}}
\hline\hline
\multirow{2}{*}{Subject ID} & \multicolumn{2}{c}{SCANimate~\cite{SCANimate:CVPR:2021}}                    & \multicolumn{2}{c}{POP~\cite{POP:ICCV:2021}}                          & \multicolumn{2}{c}{Ablation (a)}                 & \multicolumn{2}{c}{Ablation (b)}                 & \multicolumn{2}{c}{Ablation (c)}                 & \multicolumn{2}{c}{Ours, Full}                   \\ 
  \cmidrule{2-13}
     & \multicolumn{1}{c}{CD} & \multicolumn{1}{c}{NML} & \multicolumn{1}{c}{CD} & \multicolumn{1}{c}{NML} & \multicolumn{1}{c}{CD} & \multicolumn{1}{c}{NML} & \multicolumn{1}{c}{CD} & \multicolumn{1}{c}{NML} & \multicolumn{1}{c}{CD} & \multicolumn{1}{c}{NML} & \multicolumn{1}{c}{CD} & \multicolumn{1}{c}{NML} \\
\hline
anna-001      & 1.34                  & 1.35                    & 0.62                   & 0.82                    & 0.59                   & 0.82                    & 0.59                   & 0.82                    & 0.60                   & \textbf{0.81}           & \textbf{0.58}          & \textbf{0.81}           \\
beatrice-025  & 0.74                   & 1.33                    & 0.34                   & \textbf{0.75}           & 0.32                   & \textbf{0.75}           & 0.33                   & \textbf{0.75}           & 0.33                   & \textbf{0.75}           & \textbf{0.31}          & 0.77                    \\
christine-027 & 3.21                   & 1.66                    & 1.72                   & \textbf{0.97}           & 1.74                   & 1.00                    & 1.68                   & \textbf{0.97}           & 1.68                   & \textbf{0.97}           & \textbf{1.54}          & 0.99                    \\
janett-025    & 2.81                   & 1.59                    & 1.24                   & 0.89                    & 1.23                   & 0.85                    & 1.24                   & 0.82                    & 1.19                   & \textbf{0.81}           & \textbf{1.10}          & 0.82                    \\
felice-004    & 20.79                  & 2.94                   & 7.34                   & 1.24                    & 7.43                   & 1.25                    & 6.96                   & \textbf{1.22}           & 6.95                   & \textbf{1.22}           & \textbf{6.45}          & 1.25                    \\
debra-014     & 20.38                  & 2.61                    & 7.40                   & 1.26                    & 7.59                   & 1.30                    & 7.05                   & 1.28                    & 7.13                   & \textbf{1.26}           & \textbf{6.29}          & \textbf{1.26}           \\
\hline
carla-004     & 0.90                   & 1.52                    & 0.51                   & 1.02                    & 0.51                   & 1.07                    & 0.53                   & 1.06                    & 0.52                   & \textbf{1.05}           & \textbf{0.48}          & 1.06                    \\
alexandra-006 & 2.28                   & 1.84                    & 1.71                   & 1.29                    & 1.68                   & \textbf{1.28}           & 1.67                   & \textbf{1.28}           & 1.74                   & \textbf{1.28}           & \textbf{1.51}          & 1.29                    \\
eric-035      & 2.54                   & 1.94                    & 1.34                   & 1.16                    & 1.33                   & 1.16                    & 1.33                   & 1.16                    & 1.33                   & \textbf{1.13}           & \textbf{1.30}          & 1.17                    \\ 
\hline\hline
\end{tabular}
\end{table*}

\begin{figure*}[tb]
    \centering
    \includegraphics[width=\textwidth]{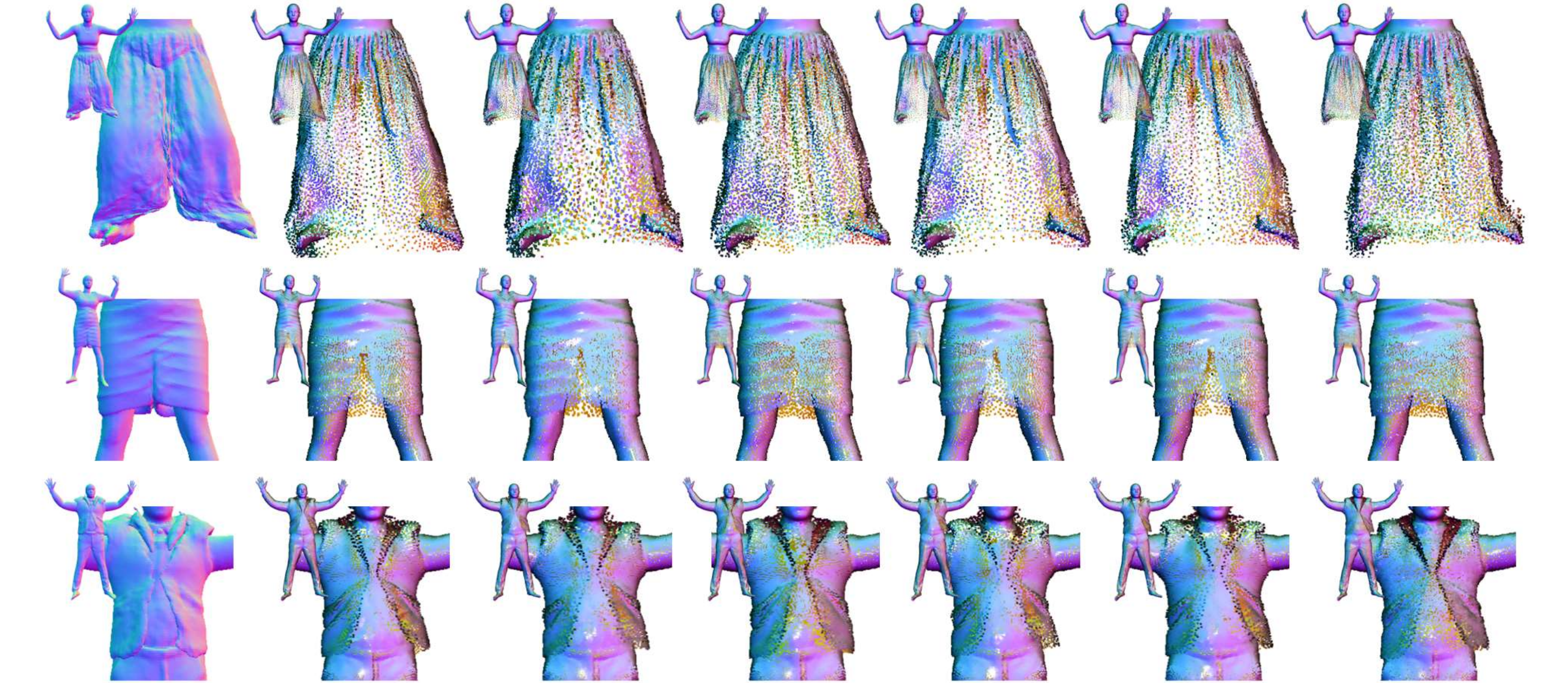}
    \put(-480, -9){\small \bf SCANimate~\cite{SCANimate:CVPR:2021} \qquad POP~\cite{POP:ICCV:2021} \quad\enspace POP + Adpt. up. \hspace{1.3em} Ablation (a) \hspace{1.4em} Ablation (b) \qquad Ablation (c) \hspace{2.2em} Full Model}
    \caption{\small Qualitative comparison with baselines and ablated versions of our model. ``Adpt. up.'' means using adaptive upsampling as a post-process.
    Subject IDs from top to bottom: ``felice-004'', ``anna-001'', ``eric-035''.
    Best viewed zoomed-in on a color screen.}
    \label{fig:qualiative_result}
\end{figure*}
\vspace{-1em}

\paragraph{Baselines.} We compare \modelname with two state-of-the-art clothed human modeling methods: SCANimate \cite{SCANimate:CVPR:2021} and POP~\cite{POP:ICCV:2021}. SCANimate uses a neural implicit shape representaiton and performs explicit data canonicalization. POP, like our method, represents clothed bodies using dense point clouds predicted in local coordinates.

We also compare with various ablated versions to our method: \textbf{(a)} we train POP with our adaptive regularization technique as described in Sec.~\ref{sec:adaptive_loss}; \textbf{(b)} we replace SMPL LBS weights in POP with our pre-diffused and smoothed LBS weights as discussed in Sec.~\ref{sec:coarse2fine}; and \textbf{(c)} we enhance POP with predicted LBS weights as described in Sec.~\ref{sec:pred_lbsw}, but without the coarse-to-fine strategy.

\label{sec:exp_setup}
\paragraph{Datasets and Metrics.} We primarily evaluate on the ReSynth dataset \cite{POP:ICCV:2021}, a synthetic dataset of clothed humans
featuring rich geometric details and salient pose-dependent clothing deformations. We pay special attention to subjects wearing skirts and dresses of varied styles, lengths and tightness, but also evaluate on several non-skirt outfits, to holistically characterize each method. Following \cite{POP:ICCV:2021}, we evaluate all the methods using the Chamfer Distance ($m^2$) and the L1 normal discrepancy, averaged over all sampled points from each method's prediction. See Appendix for more details on the evaluation setup.

\subsection{Comparison to Prior Methods} \label{sec:exp_sota_comparison}
Table~\ref{table:quant} summarizes the quantitative errors on the 9 subject-outfit types from the ReSynth dataset.
On most clothing types, our method shows a clear performance advantage over the baselines. 
As analyzed in Sec.~\ref{sec:analyze}, SCANimate often suffers from the pant-like artifacts for skirts, erroneously generating extra surfaces under the skirt and leading to high Chamfer errors especially for skirts and dresses. This shows the limitation of explicit canonicalization on these challenging clothing types. 
In contrast, POP performs reasonably well on most subjects, showing the advantage of modeling using local coordinates. 
However, the results from POP often suffer from an uneven distribution of points, especially for the long skirts and dresses, as shown in Fig.~\ref{fig:qualiative_result} (also see the teaser figure). 
Using our adaptive upsampling as post-process, the ``split'' artifact in the skirts can be partially reduced, as shown in the third column in Fig.~\ref{fig:qualiative_result}. 
In contrast, our full model achieves a more uniform point distribution for both skirts and ``plain'' clothing as shown in Fig.~\ref{fig:qualiative_result}: for skirts, it effectively eliminates the ``split'' artifact present in prior methods; for the blazer jacket it achieves the highest sharpness representing the collar and the lapel.

\subsection{Ablation Study} \label{sec:exp_ablation}
Comparing our full model with its ablated versions (a)-(c), we see the effectiveness of our introduced components.
Ablation \textbf{(a)} adds the point-adaptive regularization to POP in training, resulting in a more uniform distribution of points. However, this mechanism alone does not ensure consistently lower errors compared to the original POP model.
In ablation \textbf{(b)}, the pre-diffused LBS weight field indeed improves the numerical accuracy for all dress/skirt outfits, which, to an extent, verifies our analysis in Sec.~\ref{sec:analyze}. However, as shown in Fig.~\ref{fig:qualiative_result}, the point density still remains highly uneven for skirts. This motivates us to learn the LBS weights instead of using the fixed, pre-diffused version. 
Ablation \textbf{(c)} differs from POP in that it predicts the LBS weights. But without the coarse stage, here the predicted displacements are added directly to the body surface. 
Despite a higher accuracy than POP for all skirt outfits, qualitatively this version still suffers from an inhomogeneous point distribution especially for the long skirt. This illustrates the value of the coarse stage.
Finally, with the coarse stage, the full model achieves the lowest Chamfer error on all clothing types (both skirts and non-skirts), with an especially clear improvement on the two long skirt subjects,  ``felice-004'' and ``debra-014'', as can also be seen in the visual results. In the Appendix~\ref{sec:appendix:mesh_recon} we further discuss the influence of the point cloud quality on the mesh reconstruction.

\begin{figure}[t]
    \centering
    \includegraphics[width=0.85\linewidth]{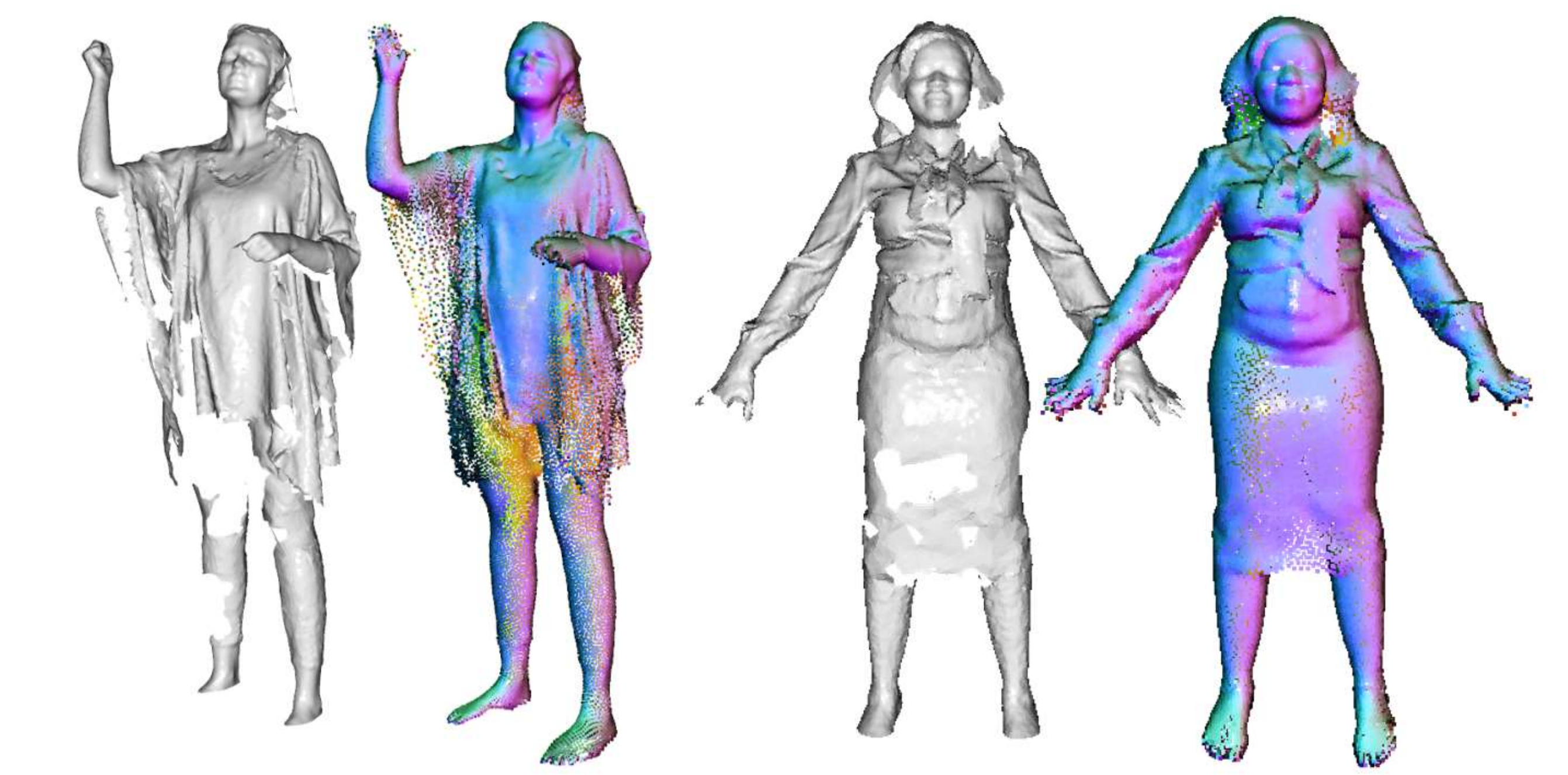}
    \caption{\small Examples of using \modelname to complete raw scans on two challenging clothing types.
    See Appendix for animated results.
    }
    \label{fig:scan_completion}
\end{figure}

\begin{figure}[t]
    \centering
    \includegraphics[width=0.9\linewidth]{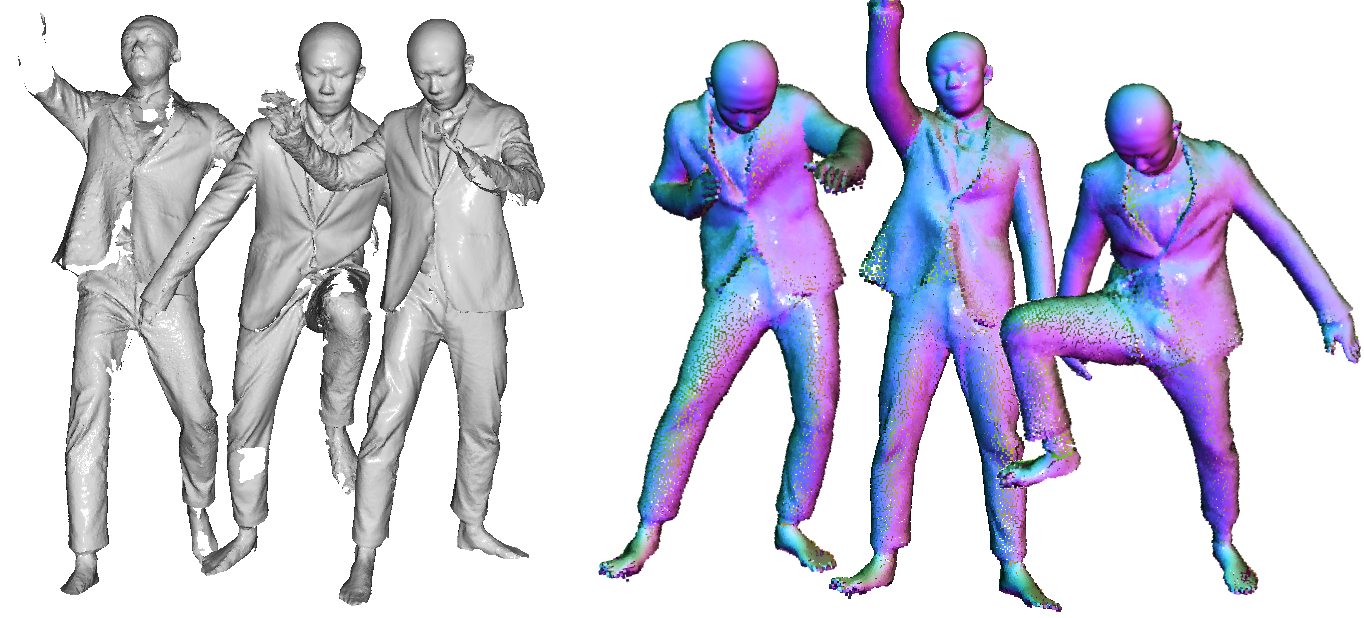}
    \caption{\small Avatar creation from raw scans. Left: raw training scan data with missing information (holes). Right: animation of learned avatar with pose-dependent clothing deformation.}
    \label{fig:raw_scan_cape}
\end{figure}

\subsection{Raw Scan Completion and Avatar Creation} \label{sec:exp:raw_scans}
\modelname learns a continuous clothing displacement field from the body, and is able to deal with raw scans too. Here we demonstrate two applications of \modelname on raw scans. Details of these experiments are provided in the Appendix.

\paragraph{Scan completion.} The raw scans from real world scanners typically contain holes, and our pipeline can serve as a solution to hole-filling and scan completion.
Consider a sequence (typically 200-300 frames) of raw, incomplete scans of a subject. We first fit the SMPL body to each frame, and then train our pipeline with the scan-body pairs. 
Our model can leverage the shared information between different data frames (i.e. the subject wears the same garment), and predict a completed shape for each frame, as shown in Fig.~\ref{fig:scan_completion}. 

\paragraph{Avatar creation from raw scans.}
As shown above, \modelname is robust to missing regions in raw scans. With sufficient data, it is possible to create a clothed human avatar with pose-dependent clothing deformation directly from raw scans. 
Here we show such an application in Fig.~\ref{fig:raw_scan_cape}. Trained directly on raw scans of the subject ``03375-blazerlong'' in the CAPE dataset~\cite{CAPE:CVPR:20}, \modelname is capable of animating the subject in novel poses with natural pose-dependent clothing deformation. 
This automated process significantly reduces the manual effort required for realistic avatar creation.
\section{Conclusion}
We have introduced a novel point-based representation for modeling clothed humans that, without loss of generality, addresses challenging clothing types such as skirts and dresses. 
It consists of three key innovations: a coarse-to-fine scheme, predicted local transformations, and point-adaptive up-sampling and regularization. 
We have systematically analyzed and evaluated our model on diverse clothing types with different skirt lengths, tightness and styles, outperforming the state-of-the-art methods and providing insights for future work on point-based representations for modeling clothed humans. 

\paragraph{Limitations and future work.}
Learning effective skinning weights for clothed body surfaces requires diverse training poses. 
When training data has very limited pose variations, our learned LBS weights (hence the local-to-world transformations) may fail to generalize to unseen poses.
The generalization in pose space with limited training data remains a challenging topic for future research. 
The point-based coarse stage can sometimes lead to slightly noisier point sets than POP; see Appendix for detailed discussions.
In this paper, we have demonstrated subject-specific modeling, but our formulation is also compatible with multi-subject training by, for example, auto-decoding a clothing geometry code for each subject as in POP~\cite{POP:ICCV:2021}. 
Given sufficient subject and clothing type variation, it should be possible to learn a shape space of clothed bodies using our approach.

\noindent
{\small 
{\bf   Acknowledgements and disclosure}:
We thank Shaofei Wang for helpful discussions. Qianli Ma is partially supported by the Max Planck ETH Center for Learning Systems.
MJB's conflicts of interest can be found at {\small \url{https://files.is.tue.mpg.de/black/CoI_3DV_2022.txt}.}
}
\clearpage
\appendix
{\noindent\Large\textbf{Appendix}}
\counterwithin{figure}{section}
\counterwithin{table}{section}

\vspace{10pt}
\noindent In this appendix we provide more details on the model designs and experimental setups (Sec.~\ref{sec:impl_details}), further elaborations on our method (Sec.~\ref{sec:method_elaboration}), and further discussions on experimental results, limitations and failure cases (Sec.~\ref{sec:further_results}). Please also refer to the video for animated results. 

\section{Implementation Details}\label{sec:impl_details}
\subsection{\modelname} \label{sec:appendix:skirt_archi}
\paragraph{Architecture.}
The coarse and the fine shape prediction have the same architectural design, but majorly differ in their inputs. Both networks are a 8-layer multi-layer perceptron (MLP) comprising 256 neurons per layer, and with the SoftPlus as the nonlinear activation. Following \cite{park2019deepsdf}, a skip connection is added from the input layer to the fourth layer of the network. From the fifth layer the network branches out two heads with the same architecture that predicts the point locations and normals respectively.

The coarse shape MLP takes as input the Cartesian coordinates of a query point from the surface of the SMPL-X body in the canonical pose $\hat{b}_i\in\mathbb{R}^3$, as well as a global shape code $z^\textrm{coarse}\in \mathbb{R}^{256}$. The global shape code is shared by all query locations for the coarse stage. 

The fine detail MLP takes a query coordinates $\hat{b}_i\in\mathbb{R}^3$, a local shape descriptor $z^\textrm{fine}_i\in \mathbb{R}^{64}$, and a local pose descriptor $z^\textrm{pose}_i\in \mathbb{R}^{64}$. Each query location on the body surface is associated with a unique local shape descriptor $z^\textrm{fine}_i$ that is an optimizable latent vector learned in an auto-decoding fashion~\cite{park2019deepsdf}. To get the pose descriptor $z^\textrm{pose}_i$, we follow~\cite{POP:ICCV:2021} and first encode the UV-positional map (with a resolution of $128\times128$) of the unclothed posed body by a U-Net~\cite{ronneberger2015u}, resulting in a $128\times128\times64$-dimensional pose feature tensor. 
For each query location $\hat{b}_i$, we find its corresponding UV-coordinate $\hat{u}_i\in \mathbb{R}^2$, and obtain the local pose feature $z^\textrm{pose}_i$ by querying the spatial dimensions (i.e. the first two dimensions) of the pose feature tensor using $\hat{u}_i$ with bilinear interpolation.
We empirically find that this U-Net-based pose feature encoder provides the sharpest geometric details. Replacing it with e.g. the filtered local pose parameters (as in SCANimate~\cite{SCANimate:CVPR:2021}) yields a significant performance drop.

The skinning weights prediction network is a 5-layer MLP that comprises 256 neurons per intermediate layer.  The final layer outputs a 22-dimentional vector that corresponds to the 22 clothing-related body joints in the SMPL-X model (we merged all finger-related joints into a ``hand'' joint for each hand).
LeakyReLU is used as the nonlinear activation except for the last layer where a SoftMax is used to obtain normalized skinning weights.

%%%%%
\paragraph{Loss functions.} 
In addition to the loss functions introduced in the main paper Eqs.~\eqref{eq:rgl}-\eqref{eq:reproj}, we follow~\cite{POP:ICCV:2021} and use Chamfer Distance and the normal loss to train our model.

The bi-directional Chamfer Distance penalizes the L2 discrepancy between the generated point cloud $\mathbf{X} = \{x_i\}$ and the ground truth $\mathbf{Y} = \{y_j\}$: $\mathcal{L}_\textrm{CD} = $
\begin{equation}
\begin{aligned}\label{eq:chamfer}
\frac{1}{|\mathbf{X}|} \sum_{i=1}^{|\mathbf{X}|} \min _{j}\norm{x_i-y_j}_2^{2} +\frac{1}{|\mathbf{Y}|}\sum_{j=1}^{|\mathbf{Y}|} \min_{i}\norm{x_i-y_j}_2^{2}.
\end{aligned}
\end{equation}

The normal loss $\mathcal{L}_\textrm{n}$ is the L1 discrepancy between each point's predicted unit normal $\bm{n}(x_i)$ and the normal of its nearest neighbor in the ground truth point cloud:
\begin{equation}\label{eq:normal_loss}
\mathcal{L}_\textrm{n} = \frac{1}{|\mathbf{X}|}\sum_{i=1}^{|\mathbf{X}|}\bignorm{\bm{n}(x_{i})
    - \bm{n}(\argmin_{y_j\in\mathbf{Y}}d(x_{i},y_j))}_1.
\end{equation}
Note that the normal loss is computed based on the nearest points found by the Chamfer Distance, 
intuitively it is more effective when the point locations of the predicted point cloud roughly matches the ground truth. Therefore we introduce the normal loss in our training after 100 epochs when the Chamfer loss plateaus.

The total loss is a weighted sum of all these terms:
\begin{equation}\label{eq:total_loss}
    \mathcal{L} = \lambda_\textrm{CD} \mathcal{L}_\textrm{CD} + \lambda_\textrm{n} \mathcal{L}_\textrm{n} + \mathcal{L}_\textrm{rgl} + \lambda_\textrm{LBS} \mathcal{L}_\textrm{LBS} + \lambda_\textrm{reproj}\mathcal{L}_\textrm{reproj}.
\end{equation}
The weights are set as $\lambda_\textrm{CD}=1e4, \lambda_n=1.0, \lambda_\textrm{LBS}=1.0,~\lambda_\textrm{reproj}=5e2$.
Note that, as discussed in the main paper, the displacement regularization term $\mathcal{L}_\textrm{rgl}$ contains a per-point adaptive weight $\lambda_i^\textrm{adapt}$ which has an initial value of $2e3$ but decays at each point adaptively according to its kNN radius. See Sec.~\ref{sec:appendix:adaptive_sample} for more details. 

%%%%%
\paragraph{Training and Inference.}
We use query points from the low-resolution ($128\times128$) UV map to train the coarse stage MLP, and evaluate it at test-time with denser query locations (e.g. those from a $256\times256$ UV map) that match the number of points on the fine stage.
Although it is also possible to train the coarse stage with denser points too, we find that the resulting coarse shape tends to be noisy.
Using fewer points for training prevents the model from being overfit to local details of certain training examples, and yields a smoother learned coarse shape. 
The coarse stage is trained using the Adam optimizer with a learning rate of $3e-4$ for 150 epochs, which takes approximately 0.9 hour on a NVIDIA RTX 6000 GPU.

For the fine stage, we train the displacement network and the LBS MLP together using Adam with a learning rate of $3e-4$ for 250 epochs, which takes approximately 2.5 hours on a NVIDIA RTX 6000 GPU.

%%%%%%%%%%%%%%%
\subsection{Baselines}
%%%%%
\paragraph{SCANimate.}
For the pose-dependent shape prediction task (main paper Sec.~\ref{sec:exp_sota_comparison}), we use the official implementation and hyperparameter settings from SCANimate~\cite{SCANimate:CVPR:2021} and train a separate model for each outfit in the ReSynth dataset~\cite{POP:ICCV:2021}.
Since SCANimate requires watertight meshes for training the implicit surfaces, we process the dense point clouds provided in the ReSynth dataset into watertight meshes using Poisson Surface Reconstruction~\cite{kazhdan2013screened} and then use this processed data as ground truth to train the model. During training, we sample 8000 points from the clothed body meshes dynamically at each iteration.

At test-time, for a fair comparison with the point-based methods, we first extract a surface from the SCANimate's implicit predictions using Marching Cubes~\cite{lorensen1987marching}, and then sample the same number (47,911) of points from the surface for evaluating the quantitative errors.

%%%%%
\paragraph{POP.} We train and evaluate POP also in a subject-specific manner to fairly compare with all other methods. 
Following the official implementation, we train and evaluate model using a fixed set of 47911 query points on the body that correspond to the valid pixels from the $256\times256$ UV-map of SMPL-X. The reported quantitative errors are averaged over all the query points.

%%%%%%%%%%%%%%%
\subsection{Further Details on Experimental Setups}
%%%%%
Here we provide more details on scan completion experiment as described in the main paper Sec.~\ref{sec:exp:raw_scans}.
We scan subjects wearing challenging clothing (e.g. skirts and very loose and wrinkly shirts) using a body scanner. The subjects are asked to improvise on several motion sequences, resulting in few hundreds of frames (typically 200-300) of raw scan data. 
We then fit the SMPL-X body model to the scan data, and train a subject-specific \modelname model with the scan-body pairs using the same procedure and hyperparameters as described in Sec.~\ref{sec:impl_details}. The subjects' hands and feet are often largely incomplete throughout the sequence in the captured raw scans due to the hardware limitations. Therefore we zero out the displacement predictions for these parts. 
At test time, we feed the model with the training bodies. The trained model is expected to reproduce the training data and outputs the hole-filled complete scans of point clouds. 

%%%%%%%%%%%%%%%%%%%%%%%%%%%%%%%%%%%%%%%%%%%%%%%%%%%%
\section{Extended Illustrations}\label{sec:method_elaboration}
\subsection{Smoothing the LBS Weight Field}\label{sec:appendix:lbs_field}
\begin{figure}
    \centering
    \includegraphics[width=\linewidth]{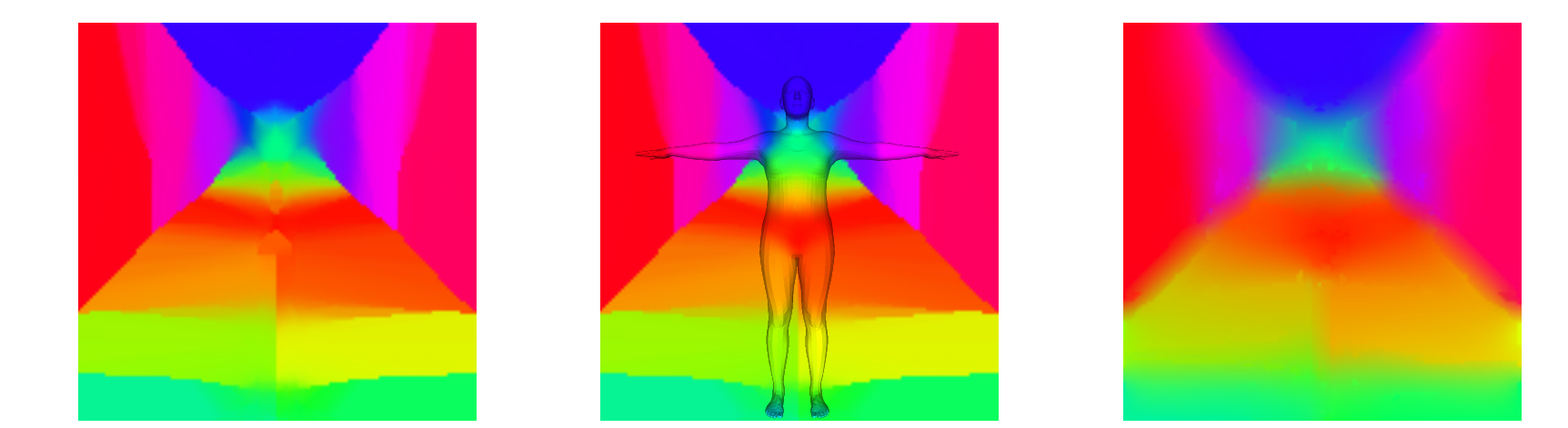}
    \put(-238, 54){(a)\hspace{6.3em} (b)\hspace{6.4em} (c)}
    \caption{Visualizing a slice from the pre-diffused SMPL-X LBS weight field. (a) Diffusion using nearest neighbor assignment. (b) Overlay of the SMPL-X body (color-coded with the body LBS weights) in the nearest-neighbor-diffused field. (c) The smoothed field after optimization convergence. The colors in the fields indicate the skinning weights to body part with the corresponding color in (b).}
    \label{fig:lbsw_field}
\end{figure}

As described in the main paper Sec.~\ref{sec:coarse2fine}, we first follow LoopReg~\cite{bhatnagar2020loopreg} and diffuse LBS weights of the SMPL-X model to $\mathbb{R}^3$ the using nearest neighbor assignment: for each query point in $\mathbb{R}^3$, we find its nearest point on the SMPL-X body surface and assign its LBS weight to the query point. In practice, the diffused LBS weights are pre-computed on a regular grid (can also be seen as voxel centers) with a resolution of $128\times128\times128$. For an arbitrary query point, its LBS weights are acquired via trilinear interpolation on its neighboring grid points. 
As shown in Fig.~\ref{fig:lbsw_field}(a), the nearest neighbor assignment results in clear discontinuities of the LBS weight distribution in space. A similar illustration can also be found in Fig.~3 in \cite{bhatnagar2020loopreg}.

To smooth the spatial distributions of the LBS weights, we use an optimization-based approach. On a high level, for each grid point, we minimize the discrepancy between its LBS weights and the average of its 1-ring neighbors. Formally, we optimize the LBS weights on all the grid points with respect to the following energy function:
\begin{equation}\label{eq:optim_smooth_lbsw}
    \mathcal{E}_\textrm{smooth} = \sum_{i=1}^M \norm{ w_{p_i} - \frac{1}{|\mathcal{N}(p_i)|}\sum_{\forall q_j \in \mathcal{N}(p_i)} w_{q_j}}_1,
\end{equation}
where $p_i$ denotes a grid point, $\mathcal{N}(p_i)$ denotes the set of its 1-ring neighbors, $q_j$ is a point in the neighborhood, and $M$ is the total number of grid points.
The optimization effectively smoothes the boundary discontinuities of the LBS weights in space, as shown in Fig.~\ref{fig:lbsw_field}(c).

\subsection{Point-adaptive Regularization/Upsampling}\label{sec:appendix:adaptive_sample}
As described in the main paper Sec.~\ref{sec:adaptive_loss}, both regularization (for training) and the point-adaptive upsampling (for post-processing) are based on the kNN radius $l_i$ , i.e. the averaged k-nearest neighbor (we used k=5) distance, per point. 
Let $\mu_l$ and $\sigma_l$ be the mean and standard deviation of points' kNN radius calculated over the entire point set. 
During training, if a point's kNN radius is above a threshold of mean plus two standard deviations, we disable the regularization on the normal of the displacement for that point:
\begin{equation}\label{eq:adaptive_weight}
    \lambda_i^\textrm{adapt} = 
    \begin{cases}
      0 & \text{if $l_i>\mu_l + 2\sigma_l$},\\
      2e3 & \text{otherwise.}\\
    \end{cases}       
\end{equation}
We also experimented with more sophisticated policy such as decaying the per-point weight using an exponential decay with respect to the kNN radius, but do not observe noticeable improvements. 

The simple thresholding in Eq.~\eqref{eq:adaptive_weight} is also applied to the upsampling in the post-process.
Consider a point $x_i$ with $l_i>\mu_l + 2\sigma_l$, we aim to generate more points around it. To do so, we find the triangle on the SMPL-X body mesh where the $x_i$'s corresponding body point $\hat{b}_i$ locates, and assign a higher sampling weight $\eta_i$ for the triangle:
\begin{equation}\label{eq:adaptive_weight}
    \eta_i^\textrm{resample} = 
    \begin{cases}
      2^{\frac{l_i - \mu}{\sigma}} & \text{if $l_i>\mu_l + 2\sigma_l$},\\
      0 & \text{otherwise.}\\
    \end{cases}       
\end{equation}
We then re-sample query points from the body mesh surface by modifying the standard uniform mesh sampling. In the uniform sampling, the sampling probability on each triangle is proportional to its area. We scale this probability with $\eta_i^\textrm{resample}$, so that more points are generated on the regions that originally have lower point density. 
The resampled query points are then fed into the MLP to generate the new points on the clothing surface.
The final result is the union of the new points and the original point set. 
This process can be repeated for several iterations for improved visual quality. The results shown in the main paper (third column in Fig.~\ref{fig:qualiative_result}) uses 3 iterations of upsampling. 

%%%%%%%%%%%%%%%%%%%%%%%%%%%%%%%%%%%%%%%%%%%%%%%%%%%%
\section{Extended Results and Discussions}\label{sec:further_results}
\subsection{Limitation and Failure Cases}
In the case of very loose dresses, the points on the dress surface produced by \modelname can still be visibly sparser than on other body parts as visualized in the main paper Fig.~\ref{fig:qualiative_result}; consequently the mesh reconstruction can have rough surfaces as shown in Fig.~\ref{fig:meshing_comparison}.

Although the coarse stage enables handling challenging clothing types, the point-based coarse geometry is typically not as smooth as that of the base body mesh. Consequently, when adding the fine-stage predictions to the coarse shape, the final geometry can possibly be noisier than POP (which adds displacements directly to the unclothed body), hence occasional loss of clothing details (as seen in the main paper Fig.~\ref{fig:qualiative_result}) and a noisier surface reconstruction (as shown in Fig.~\ref{fig:meshing_comparison}), despite a higher quantitative accuracy on entire test set.
Unlike the mesh representation where one could apply loss functions (such as the Laplacian term) that encourage the smoothness in the generated geometry, for point clouds such regularization is not trivial due to the absence of the point connectivity, and we leave this for future work.

As with other recent models~\cite{SCANimate:CVPR:2021, POP:ICCV:2021}, our model requires the fitted underlying body to be accurate. While this is not an issue with the synthetic ReSynth data that we use in the paper, we observe certain failure cases with the real data where the estimated body under clothing is imperfect.
For example, in the scan completion experiment, the fitted body can be inaccurate due to the challenging clothing. Consequently, in the resulting completed scans flickering can be observed especially at the extremities, as shown in the supplementary video.
In the future we plan to refine the body pose together with the network parameters during training so as to better handle real world data. 

\subsection{Mesh Reconstruction} \label{sec:appendix:mesh_recon}
\begin{figure*}[tb]
    \centering
    \includegraphics[width=\textwidth]{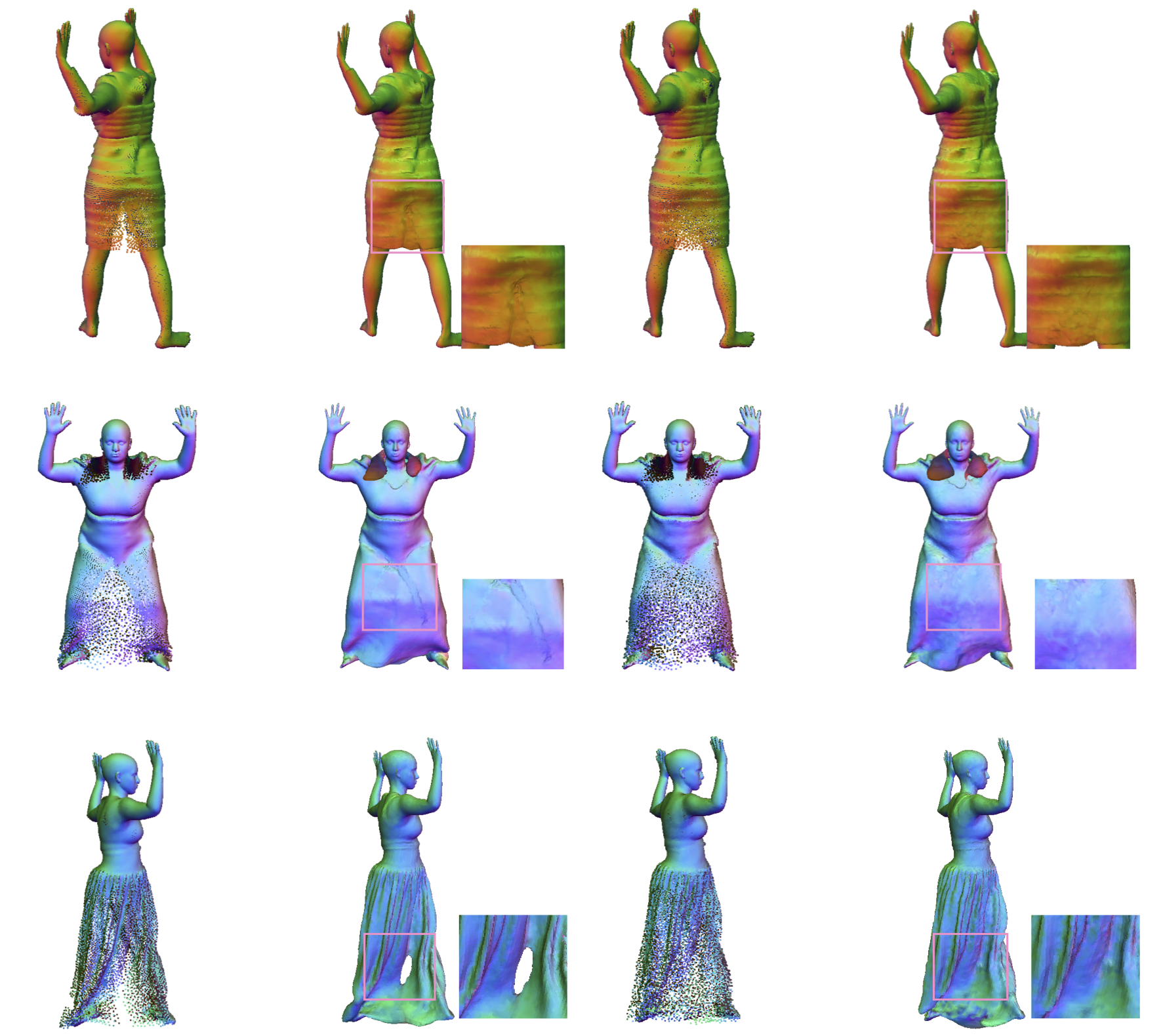}
    \put(-465,455){POP~\cite{POP:ICCV:2021} \hspace{7.5em} POP, meshed \hspace{7.5em} Ours \hspace{9em} Ours, meshed}
    \caption{Comparison between the point-based baseline POP~\cite{POP:ICCV:2021} and our method in terms of mesh reconstruction quality. The meshed results are produced by performing Screened Poisson Surface Reconstruction~\cite{kazhdan2013screened} on the corresponding point clouds. 
    When the clothed body point cloud has clear gaps on the clothing surface, the reconstructed mesh is prone to artifacts.}
    \label{fig:meshing_comparison}
\end{figure*}

In Fig.~\ref{fig:meshing_comparison} we qualitatively show the results of applying Poisson Surface Reconstruction (PSR) to the point sets generated by a baseline method and ours, respectively. 
Through building an implicit field of the indicator function, the PSR is capable of filling holes in the point clouds to a certain extent. However, when the point cloud has obvious discontinuities as with the previous method \cite{POP:ICCV:2021}, the reconstructed meshes are prone to artifacts. In contrast, our method generates point clouds that have a more uniform distribution of points on the clothing surface, and thus effectively alleviates this issue.

Note that the purpose of this experiment is to show the impact of the split-like artifacts on the quality of mesh reconstruction. 
However, mesh reconstruction is not the sole exit of point-based human representations. 
Recent work has shown the potential of combining the point-based representation with neural rendering to achieve realistic renderings without the intermediate meshing step~\cite{SCALE:CVPR:2021, prokudin2020smplpix,aliev2019neural}.
Combining these methods with our pipeline is an promising direction for future research.
\subsection{Scan Completion}
Please refer to the supplementary video for the animated results of the scan completion experiment.

{\small
\bibliographystyle{ieee_fullname}
\bibliography{references}

\begin{thebibliography}{10}\itemsep=-1pt

\bibitem{aliev2019neural}
Kara-Ali Aliev, Artem Sevastopolsky, Maria Kolos, Dmitry Ulyanov, and Victor
  Lempitsky.
\newblock Neural point-based graphics.
\newblock In {\em Proceedings of the European Conference on Computer Vision
  ({ECCV})}, pages 696--712, 2020.

\bibitem{anguelov2005scape}
Dragomir Anguelov, Praveen Srinivasan, Daphne Koller, Sebastian Thrun, Jim
  Rodgers, and James Davis.
\newblock {SCAPE}: shape completion and animation of people.
\newblock In {\em ACM Transactions on Graphics (TOG)}, volume~24, pages
  408--416, 2005.

\bibitem{bertiche2020cloth3d}
Hugo Bertiche, Meysam Madadi, and Sergio Escalera.
\newblock {CLOTH3D}: clothed 3d humans.
\newblock In {\em European Conference on Computer Vision}, pages 344--359.
  Springer, 2020.

\bibitem{bhatnagar2020ipnet}
Bharat~Lal Bhatnagar, Cristian Sminchisescu, Christian Theobalt, and Gerard
  Pons-Moll.
\newblock Combining implicit function learning and parametric models for {3D}
  human reconstruction.
\newblock In {\em Proceedings of the European Conference on Computer Vision
  ({ECCV})}, pages 311--329, 2020.

\bibitem{bhatnagar2020loopreg}
Bharat~Lal Bhatnagar, Cristian Sminchisescu, Christian Theobalt, and Gerard
  Pons-Moll.
\newblock {LoopReg}: Self-supervised learning of implicit surface
  correspondences, pose and shape for {3D} human mesh registration.
\newblock In {\em Advances in Neural Information Processing Systems
  ({NeurIPS})}, pages 12909--12922, 2020.

\bibitem{burov2021dsfn}
Andrei Burov, Matthias Nie{\ss}ner, and Justus Thies.
\newblock Dynamic surface function networks for clothed human bodies.
\newblock In {\em Proceedings of the IEEE/CVF International Conference on
  Computer Vision (ICCV)}, 2021.

\bibitem{xu2022gdna}
Xu Chen, Tianjian Jiang, Jie Song, Jinlong Yang, Michael~J. Black, Andreas
  Geiger, and Otmar Hilliges.
\newblock {gDNA}: Towards generative detailed neural avatars.
\newblock In {\em IEEE/CVF Conf.~on Computer Vision and Pattern Recognition
  (CVPR)}, pages 20427--20437, June 2022.

\bibitem{chen2021snarf}
Xu Chen, Yufeng Zheng, Michael~J Black, Otmar Hilliges, and Andreas Geiger.
\newblock {SNARF}: Differentiable forward skinning for animating non-rigid
  neural implicit shapes.
\newblock In {\em Proceedings of the IEEE/CVF International Conference on
  Computer Vision (ICCV)}, 2021.

\bibitem{chen2019imnet}
Zhiqin Chen and Hao Zhang.
\newblock Learning implicit fields for generative shape modeling.
\newblock In {\em Proceedings IEEE Conf. on Computer Vision and Pattern
  Recognition (CVPR)}, pages 5939--5948, 2019.

\bibitem{chibane2020implicit}
Julian Chibane, Thiemo Alldieck, and Gerard Pons-Moll.
\newblock Implicit functions in feature space for {3D} shape reconstruction and
  completion.
\newblock In {\em Proceedings IEEE Conf. on Computer Vision and Pattern
  Recognition (CVPR)}, pages 6970--6981, 2020.

\bibitem{chibane2020ndf}
Julian Chibane, Aymen Mir, and Gerard Pons-Moll.
\newblock Neural unsigned distance fields for implicit function learning.
\newblock In {\em Advances in Neural Information Processing Systems
  ({NeurIPS})}, pages 21638--21652, 2020.

\bibitem{de2010stable}
Edilson De~Aguiar, Leonid Sigal, Adrien Treuille, and Jessica~K Hodgins.
\newblock Stable spaces for real-time clothing.
\newblock In {\em ACM Transactions on Graphics (TOG)}, volume~29, page 106.
  ACM, 2010.

\bibitem{dong2022pina}
Zijian Dong, Chen Guo, Jie Song, Xu Chen, Andreas Geiger, and Otmar Hilliges.
\newblock {PINA}: Learning a personalized implicit neural avatar from a single
  {RGB-D} video sequence.
\newblock In {\em Proceedings IEEE Conf. on Computer Vision and Pattern
  Recognition (CVPR)}, pages 20470--20480, 2022.

\bibitem{gong2019spiralnet++}
Shunwang Gong, Lei Chen, Michael Bronstein, and Stefanos Zafeiriou.
\newblock Spiralnet++: A fast and highly efficient mesh convolution operator.
\newblock In {\em Proceedings of the IEEE/CVF International Conference on
  Computer Vision Workshops}, pages 0--0, 2019.

\bibitem{guan2012drape}
Peng Guan, Loretta Reiss, David~A Hirshberg, Alexander Weiss, and Michael~J
  Black.
\newblock {DRAPE: DRessing Any PErson.}
\newblock {\em ACM Transactions on Graphics (TOG)}, 31(4):35--1, 2012.

\bibitem{habermann2020deepcap}
Marc Habermann, Weipeng Xu, Michael Zollhofer, Gerard Pons-Moll, and Christian
  Theobalt.
\newblock Deepcap: Monocular human performance capture using weak supervision.
\newblock In {\em Proceedings of the IEEE/CVF Conference on Computer Vision and
  Pattern Recognition}, pages 5052--5063, 2020.

\bibitem{jiang2020bcnet}
Boyi Jiang, Juyong Zhang, Yang Hong, Jinhao Luo, Ligang Liu, and Hujun Bao.
\newblock {BCNet}: Learning body and cloth shape from a single image.
\newblock In {\em Proceedings of the European Conference on Computer Vision
  ({ECCV})}, pages 18--35. Springer, 2020.

\bibitem{joo2018total}
Hanbyul Joo, Tomas Simon, and Yaser Sheikh.
\newblock {Total Capture}: A {3D} deformation model for tracking faces, hands,
  and bodies.
\newblock In {\em Proceedings IEEE Conf. on Computer Vision and Pattern
  Recognition (CVPR)}, pages 8320--8329, 2018.

\bibitem{kazhdan2013screened}
Michael Kazhdan and Hugues Hoppe.
\newblock Screened {Poisson} surface reconstruction.
\newblock {\em ACM Transactions on Graphics (TOG)}, 32(3):1--13, 2013.

\bibitem{lahner2018deepwrinkles}
Zorah L{\"a}hner, Daniel Cremers, and Tony Tung.
\newblock {DeepWrinkles}: Accurate and realistic clothing modeling.
\newblock In {\em Proceedings of the European Conference on Computer Vision
  ({ECCV})}, pages 698--715, 2018.

\bibitem{lin2022fite}
Siyou Lin, Hongwen Zhang, Zerong Zheng, Ruizhi Shao, and Yebin Liu.
\newblock Learning implicit templates for point-based clothed human modeling.
\newblock In {\em Proceedings of the European Conference on Computer Vision
  ({ECCV})}, 2022.

\bibitem{liu2019neuroskinning}
Lijuan Liu, Youyi Zheng, Di Tang, Yi Yuan, Changjie Fan, and Kun Zhou.
\newblock {NeuroSkinning}: Automatic skin binding for production characters
  with deep graph networks.
\newblock {\em ACM Transactions on Graphics (TOG)}, 38(4):1--12, 2019.

\bibitem{loper2015smpl}
Matthew Loper, Naureen Mahmood, Javier Romero, Gerard Pons-Moll, and Michael~J
  Black.
\newblock {SMPL}: A skinned multi-person linear model.
\newblock {\em ACM Transactions on Graphics (TOG)}, 34(6):248, 2015.

\bibitem{lorensen1987marching}
William~E Lorensen and Harvey~E Cline.
\newblock Marching cubes: A high resolution {3D} surface construction
  algorithm.
\newblock In {\em ACM {SIGGRAPH} Computer Graphics}, volume~21, pages 163--169,
  1987.

\bibitem{SCALE:CVPR:2021}
Qianli Ma, Shunsuke Saito, Jinlong Yang, Siyu Tang, and Michael~J. Black.
\newblock {SCALE}: Modeling clothed humans with a surface codec of articulated
  local elements.
\newblock In {\em Proceedings IEEE/CVF Conf.~on Computer Vision and Pattern
  Recognition (CVPR)}, pages 16082--16093, June 2021.

\bibitem{CAPE:CVPR:20}
Qianli Ma, Jinlong Yang, Anurag Ranjan, Sergi Pujades, Gerard Pons-Moll, Siyu
  Tang, and Michael~J. Black.
\newblock Learning to dress {3D} people in generative clothing.
\newblock In {\em Proceedings IEEE Conf. on Computer Vision and Pattern
  Recognition (CVPR)}, pages 6468--6477, 2020.

\bibitem{POP:ICCV:2021}
Qianli Ma, Jinlong Yang, Siyu Tang, and Michael~J. Black.
\newblock The power of points for modeling humans in clothing.
\newblock In {\em Proceedings of the IEEE/CVF International Conference on
  Computer Vision (ICCV)}, pages 10974--10984, Oct. 2021.

\bibitem{mescheder2019occupancy}
Lars Mescheder, Michael Oechsle, Michael Niemeyer, Sebastian Nowozin, and
  Andreas Geiger.
\newblock Occupancy networks: Learning {3D} reconstruction in function space.
\newblock In {\em Proceedings IEEE Conf. on Computer Vision and Pattern
  Recognition (CVPR)}, pages 4460--4470, 2019.

\bibitem{Neophytou2014layered}
Alexandros Neophytou and Adrian Hilton.
\newblock A layered model of human body and garment deformation.
\newblock In {\em International Conference on {3D} Vision (3DV)}, pages
  171--178, 2014.

\bibitem{palafox2021npm}
Pablo Palafox, Aljaz Bozic, Justus Thies, Matthias Nie{\ss}ner, and Angela Dai.
\newblock Neural parametric models for {3D} deformable shapes.
\newblock In {\em Proceedings of the IEEE/CVF International Conference on
  Computer Vision (ICCV)}, 2021.

\bibitem{palafox2021spams}
Pablo Palafox, Nikolaos Sarafianos, Tony Tung, and Angela Dai.
\newblock Spams: Structured implicit parametric models.
\newblock {\em CVPR}, 2022.

\bibitem{pan2022predicting}
Xiaoyu Pan, Jiaming Mai, Xinwei Jiang, Dongxue Tang, Jingxiang Li, Tianjia
  Shao, Kun Zhou, Xiaogang Jin, and Dinesh Manocha.
\newblock Predicting loose-fitting garment deformations using bone-driven
  motion networks.
\newblock {\em SIGGRAPH}, 2022.

\bibitem{park2019deepsdf}
Jeong~Joon Park, Peter Florence, Julian Straub, Richard Newcombe, and Steven
  Lovegrove.
\newblock {DeepSDF}: Learning continuous signed distance functions for shape
  representation.
\newblock In {\em Proceedings IEEE Conf. on Computer Vision and Pattern
  Recognition (CVPR)}, pages 165--174, 2019.

\bibitem{patel20tailornet}
Chaitanya Patel, Zhouyingcheng Liao, and Gerard Pons-Moll.
\newblock {TailorNet}: Predicting clothing in {3D} as a function of human pose,
  shape and garment style.
\newblock In {\em Proceedings IEEE Conf. on Computer Vision and Pattern
  Recognition (CVPR)}, pages 7363--7373, 2020.

\bibitem{AGORA:CVPR:21}
Priyanka Patel, Chun-Hao~Huang Paul, Joachim Tesch, David Hoffmann, Shashank
  Tripathi, and Michael~J. Black.
\newblock {AGORA}: Avatars in geography optimized for regression analysis.
\newblock In {\em Proceedings IEEE Conf. on Computer Vision and Pattern
  Recognition (CVPR)}, pages 13468--13478, June 2021.

\bibitem{pavlakos2019expressive}
Georgios Pavlakos, Vasileios Choutas, Nima Ghorbani, Timo Bolkart, Ahmed A.~A.
  Osman, Dimitrios Tzionas, and Michael~J. Black.
\newblock Expressive body capture: {3D} hands, face, and body from a single
  image.
\newblock In {\em Proceedings IEEE Conf. on Computer Vision and Pattern
  Recognition (CVPR)}, pages 10975--10985, 2019.

\bibitem{pons2017clothcap}
Gerard Pons-Moll, Sergi Pujades, Sonny Hu, and Michael~J Black.
\newblock {ClothCap}: Seamless {4D} clothing capture and retargeting.
\newblock {\em ACM Transactions on Graphics (TOG)}, 36(4):73, 2017.

\bibitem{prokudin2020smplpix}
Sergey Prokudin, Michael~J. Black, and Javier Romero.
\newblock {SMPLpix}: Neural avatars from {3D} human models.
\newblock In {\em Winter Conference on Applications of Computer Vision (WACV)},
  2021.

\bibitem{ranjan2018generatingcoma}
Anurag Ranjan, Timo Bolkart, Soubhik Sanyal, and Michael~J Black.
\newblock Generating {3D} faces using convolutional mesh autoencoders.
\newblock In {\em Proceedings of the European Conference on Computer Vision
  ({ECCV})}, pages 725--741, 2018.

\bibitem{ronneberger2015u}
Olaf Ronneberger, Philipp Fischer, and Thomas Brox.
\newblock {U-Net}: Convolutional networks for biomedical image segmentation.
\newblock In {\em International Conference on Medical Image Computing and
  Computer-Assisted Intervention (MICCAI)}, pages 234--241, 2015.

\bibitem{saito2019pifu}
Shunsuke Saito, Zeng Huang, Ryota Natsume, Shigeo Morishima, Angjoo Kanazawa,
  and Hao Li.
\newblock {PIFu}: Pixel-aligned implicit function for high-resolution clothed
  human digitization.
\newblock In {\em Proceedings of the IEEE/CVF International Conference on
  Computer Vision (ICCV)}, pages 2304--2314, 2019.

\bibitem{saito2020pifuhd}
Shunsuke Saito, Tomas Simon, Jason Saragih, and Hanbyul Joo.
\newblock {PIFuHD}: Multi-level pixel-aligned implicit function for
  high-resolution {3D} human digitization.
\newblock In {\em Proceedings IEEE Conf. on Computer Vision and Pattern
  Recognition (CVPR)}, pages 84--93, 2020.

\bibitem{SCANimate:CVPR:2021}
Shunsuke Saito, Jinlong Yang, Qianli Ma, and Michael~J. Black.
\newblock {SCANimate}: Weakly supervised learning of skinned clothed avatar
  networks.
\newblock In {\em Proceedings IEEE/CVF Conf.~on Computer Vision and Pattern
  Recognition (CVPR)}, June 2021.

\bibitem{santesteban2019}
Igor Santesteban, Miguel~A. Otaduy, and Dan Casas.
\newblock {Learning-Based Animation of Clothing for Virtual Try-On}.
\newblock {\em Computer Graphics Forum}, 38(2):355--366, 2019.

\bibitem{santesteban2022snug}
Igor Santesteban, Miguel~A Otaduy, and Dan Casas.
\newblock {SNUG}: Self-supervised neural dynamic garments.
\newblock In {\em Proceedings IEEE Conf. on Computer Vision and Pattern
  Recognition (CVPR)}, pages 8140--8150, 2022.

\bibitem{tiwari20sizer}
Garvita Tiwari, Bharat~Lal Bhatnagar, Tony Tung, and Gerard Pons-Moll.
\newblock {SIZER}: A dataset and model for parsing {3D} clothing and learning
  size sensitive {3D} clothing.
\newblock In {\em Proceedings of the European Conference on Computer Vision
  ({ECCV})}, volume 12348, pages 1--18, 2020.

\bibitem{tiwari21neuralgif}
Garvita Tiwari, Nikolaos Sarafianos, Tony Tung, and Gerard Pons-Moll.
\newblock Neural-gif: Neural generalized implicit functions for animating
  people in clothing.
\newblock In {\em International Conference on Computer Vision ({ICCV})},
  October 2021.

\bibitem{verma2018feastnet}
Nitika Verma, Edmond Boyer, and Jakob Verbeek.
\newblock Feastnet: Feature-steered graph convolutions for 3d shape analysis.
\newblock In {\em Proceedings of the IEEE conference on computer vision and
  pattern recognition}, pages 2598--2606, 2018.

\bibitem{vidaurre2020fully}
Raquel Vidaurre, Igor Santesteban, Elena Garces, and Dan Casas.
\newblock Fully convolutional graph neural networks for parametric virtual
  try-on.
\newblock In {\em Computer Graphics Forum}, volume~39, pages 145--156. Wiley
  Online Library, 2020.

\bibitem{vlasic2008articulated}
Daniel Vlasic, Ilya Baran, Wojciech Matusik, and Jovan Popovi{\'c}.
\newblock Articulated mesh animation from multi-view silhouettes.
\newblock In {\em ACM SIGGRAPH 2008 papers}, pages 1--9. 2008.

\bibitem{PTF:CVPR:2021}
Shaofei Wang, Andreas Geiger, and Siyu Tang.
\newblock Locally aware piecewise transformation fields for {3D} human mesh
  registration.
\newblock In {\em Proceedings of the IEEE/CVF Conference on Computer Vision and
  Pattern Recognition (CVPR)}, pages 7639--7648, June 2021.

\bibitem{MetaAvatar:arXiv:21}
Shaofei Wang, Marko Mihajlovic, Qianli Ma, Andreas Geiger, and Siyu Tang.
\newblock {MetaAvatar}: Learning animatable clothed human models from few depth
  images.
\newblock In {\em Advances in Neural Information Processing Systems
  ({NeurIPS})}, 2021.

\bibitem{ARAH:ECCV:2022}
Shaofei Wang, Katja Schwarz, Andreas Geiger, and Siyu Tang.
\newblock Arah: Animatable volume rendering of articulated human sdfs.
\newblock In {\em Proceedings of the European Conference on Computer Vision
  ({ECCV})}, 2022.

\bibitem{xiu2022icon}
Yuliang Xiu, Jinlong Yang, Dimitrios Tzionas, and Michael~J. Black.
\newblock {ICON}: {I}mplicit {C}lothed humans {O}btained from {N}ormals.
\newblock In {\em IEEE/CVF Conf.~on Computer Vision and Pattern Recognition
  (CVPR)}, pages 13296--13306, June 2022.

\bibitem{xu2021hnerf}
Hongyi Xu, Thiemo Alldieck, and Cristian Sminchisescu.
\newblock {H-NeRF}: Neural radiance fields for rendering and temporal
  reconstruction of humans in motion.
\newblock {\em Advances in Neural Information Processing Systems ({NeurIPS})},
  34:14955--14966, 2021.

\bibitem{xu2020ghum}
Hongyi Xu, Eduard~Gabriel Bazavan, Andrei Zanfir, William~T Freeman, Rahul
  Sukthankar, and Cristian Sminchisescu.
\newblock {GHUM \& GHUML}: Generative {3D} human shape and articulated pose
  models.
\newblock In {\em Proceedings IEEE Conf. on Computer Vision and Pattern
  Recognition (CVPR)}, pages 6184--6193, 2020.

\bibitem{xu2014sensitivity}
Weiwei Xu, Nobuyuki Umetani, Qianwen Chao, Jie Mao, Xiaogang Jin, and Xin Tong.
\newblock Sensitivity-optimized rigging for example-based real-time clothing
  synthesis.
\newblock {\em ACM Trans. Graph.}, 33(4):107--1, 2014.

\bibitem{yang2018analyzing}
Jinlong Yang, Jean-S{\'e}bastien Franco, Franck H{\'e}troy-Wheeler, and
  Stefanie Wuhrer.
\newblock Analyzing clothing layer deformation statistics of {3D} human
  motions.
\newblock In {\em Proceedings of the European Conference on Computer Vision
  ({ECCV})}, pages 237--253, 2018.

\bibitem{Zhang_2017_CVPR}
Chao Zhang, Sergi Pujades, Michael~J. Black, and Gerard Pons-Moll.
\newblock Detailed, accurate, human shape estimation from clothed 3d scan
  sequences.
\newblock In {\em The IEEE Conference on Computer Vision and Pattern
  Recognition (CVPR)}, July 2017.

\bibitem{zhu2020deep}
Heming Zhu, Yu Cao, Hang Jin, Weikai Chen, Dong Du, Zhangye Wang, Shuguang Cui,
  and Xiaoguang Han.
\newblock {Deep Fashion3D}: A dataset and benchmark for {3D} garment
  reconstruction from single images.
\newblock In {\em Proceedings of the European Conference on Computer Vision
  ({ECCV})}, volume 12346, pages 512--530, 2020.

\bibitem{zhu2022registering}
Heming Zhu, Lingteng Qiu, Yuda Qiu, and Xiaoguang Han.
\newblock Registering explicit to implicit: {T}owards high-fidelity garment
  mesh reconstruction from single images.
\newblock In {\em Proceedings IEEE Conf. on Computer Vision and Pattern
  Recognition (CVPR)}, pages 3845--3854, June 2022.

\end{thebibliography}
}

\end{document}